\documentclass{article}

\usepackage{arxiv}

\usepackage{amsmath,amsfonts,bm}

\def\eqref#1{equation~\ref{#1}}

\def\1{\bm{1}}

\DeclareMathAlphabet{\mathsfit}{\encodingdefault}{\sfdefault}{m}{sl}
\SetMathAlphabet{\mathsfit}{bold}{\encodingdefault}{\sfdefault}{bx}{n}

\usepackage{hyperref}
\usepackage{url}
\usepackage[many]{tcolorbox}
\usepackage{subcaption}
\usepackage{algorithm}
\usepackage{algpseudocode}
\usepackage{pifont}
\usepackage{booktabs}
\usepackage{graphicx}
\usepackage{multirow}
\usepackage{subcaption}
\usepackage[normalem]{ulem}
\useunder{\uline}{\ul}{}
\usepackage{mathtools}
\usepackage[table]{xcolor}

\usepackage{amsthm}

\newtheorem{proposition}{Proposition}
\newtheorem{lemma}{Lemma}

\theoremstyle{definition}

\theoremstyle{remark}

\usepackage{wrapfig}
\usepackage{enumitem}
\usepackage{authblk}

\definecolor{darkgrey}{rgb}{0.53,0.53,0.53}
\definecolor{mygrey}{rgb}{0.9,0.9,0.9}
\newcommand{\vpara}[1]{\vspace{0.00in}\noindent\textbf{#1 }}

\usepackage{natbib}

\title{\texttt{OATS}: Online Data Augmentation for Time Series Foundation Models}

\date{} 					

\author[1]{Junwei Deng}
\author[2,*]{Chang Xu}
\author[1]{Jiaqi W. Ma}
\author[3]{Ming Jin}
\author[4]{Chenghao Liu}
\author[2]{Xu Zhang}
\author[2]{Li Zhao}
\author[2]{Jiang Bian}

\affil[1]{University of Illinois Urbana-Champaign}
\affil[2]{Microsoft Research}
\affil[3]{Griffith University}
\affil[4]{Datadog AI Research}
\affil[*]{Corresponding author. Email: \texttt{chanx@microsoft.com}}

\renewcommand{\headeright}{}
\renewcommand{\undertitle}{}
\renewcommand{\shorttitle}{\texttt{OATS}: Online Data Augmentation for Time Series Foundation Models}

\hypersetup{
pdftitle={\texttt{OATS}: Online Data Augmentation for Time Series Foundation Models},
pdfsubject={cs.LG},
pdfauthor={Junwei~Deng, Chang~Xu, Jiaqi~Ma, Ming~Jin, Jiang~Bian},
pdfkeywords={time series foundation model, data augmentation},
}

\begin{document}
\maketitle

\begin{abstract}
Time Series Foundation Models (TSFMs) are a powerful paradigm for time series analysis and are often enhanced by synthetic data augmentation to improve the training data quality. Existing augmentation methods, however, typically rely on heuristics and static paradigms. Motivated by dynamic data optimization, which shows that the contribution of samples varies across training stages, we propose \texttt{OATS} (\underline{O}nline Data \underline{A}ugmentation for \underline{T}ime \underline{S}eries Foundation Models), a principled strategy that generates synthetic data tailored to different training steps. \texttt{OATS} leverages valuable training samples as principled guiding signals and dynamically generates high-quality synthetic data conditioned on them. We further design a diffusion-based framework to produce realistic time series and introduce an explore-exploit mechanism to balance efficiency and effectiveness. Experiments on TSFMs demonstrate that \texttt{OATS} consistently outperforms regular training and yields substantial performance gains over static data augmentation baselines across six datasets and two TSFM architectures. The code is available at the link \url{https://github.com/microsoft/TimeCraft}.
\end{abstract}

\keywords{time series foundation model, data augmentation}

\section{Introduction}

Time series modeling plays a critical role across a wide range of domains, including finance~\citep{kim2019hats,li2024mars}, healthcare~\citep{guo2023ehr}, climate science~\citep{liang2023airformer}, and industrial monitoring~\citep{zamanzadeh2024deep}.
Recent developments in time series foundation models (TSFMs) have further advanced this field by leveraging large-scale datasets collected from multiple third-party sources~\citep{yao2024towards,ansari2024chronos}, enabling cross-domain learning and zero-shot generalization.
Nevertheless, the success of TSFMs relies on the availability of high-quality data. 
Typical challenges in time series datasets include missing values~\citep{junninen2004methods}, heterogeneous sampling rates~\citep{woo2024unified}, imbalanced domain distributions~\citep{yao2024towards}, data duplication~\citep{lin2023common}.
These issues make it more difficult to curate reliable large-scale time series datasets than in domains such as natural language processing~\citep{gao2020pile,raffel2020exploring}. As the community moves toward large-scale foundation models, synthetic data augmentation has emerged as a critical and widely adopted method for enhancing training data with synthetic samples. Leveraging the high controllability of time series patterns and operational simplicity, synthetic data augmentation can not only address data scarcity but also enriches domain diversity and improves the robustness of TSFMs~\citep{liu2025empowering}.

Various data augmentation methods have been proposed in TSFM studies to enrich training datasets with realistic and diverse synthetic data. These approaches can be broadly categorized into two groups. The first involves directly introducing \emph{manually designed patterns} to synthesize data. Such patterns include sinusoidal waves~\citep{goswami2024moment}, decomposed time series~\citep{10.5555/3666122.3666234,lan2025foundationmodelszeroshottime}, and handcrafted kernel banks~\citep{shi2024time, ansari2024chronos,xie2025caukerclassificationtimeseries}. The second group focuses on applying \emph{basic transformations to existing time series data} to generate synthetic variants. Examples of such techniques include smoothing, jittering~\citep{um2017data,moroshan2025tempopfnsyntheticpretraininglinear}, and TSMixup~\citep{ansari2024chronos,ansari2025chronos2univariateuniversalforecasting,moroshan2025tempopfnsyntheticpretraininglinear}.

While these approaches have shown empirical effectiveness, they rely on \emph{handcrafted heuristics} that are often \emph{agnostic to the model training process}, which result in two critical challenges in designing high-performance data augmentation strategies. First, the quality of time series data for TSFMs is difficult to quantify in a principled way. The heuristics that work for one time series task may fail for another~\citep{liu2025empowering,kuvshinova2024towards}. Second, recent studies show that the value of the same data sample varies across the training process~\citep{wang2024data}, which challenges the capacity of the static data augmentation paradigm that generates synthetic data once and uniformly incorporated along the whole training process.

In this study, we address these challenges by leveraging recent advancements in \emph{data attribution}~\citep{koh2017understanding,deng2025survey} and \emph{online data optimization}~\citep{wang2024greats}. For the first challenge, data attribution aims to quantify the influence of individual training data points on model outputs; instead of studying heuristic rules like which pattern of time series data is helpful for the training, data attribution allows us to define and assess the quality of time series data by its influence on a utility function, e.g., loss on a reference set, in a principle way. For the second challenge, we go beyond the static data augmentation paradigm in existing TSFM literature and conduct online data augmentation, which incorporates the training process information and dynamically generates high-quality data for each training step.

Concretely, we propose \texttt{OATS} (\underline{O}nline Data \underline{A}ugmentation for \underline{T}ime \underline{S}eries Foundation Models), a strategy to dynamically generate high-quality synthetic data in a principled manner. \texttt{OATS} generates high-quality synthetic data by using training samples with high data attribution scores~\citep{koh2017understanding} as guiding signals. \texttt{OATS} consists of three core components: \textbf{Time-series influence scores (TSIS)} integrate data attribution with time series–specific knowledge to dynamically assess the quality of each training sample in a principled manner to create a generation guiding signal. \textbf{High-quality guided data augmentation} leverages the guiding signal to condition a diffusion model trained on a small subset of the TSFM training data for synthetic data generation. To reduce computational overhead and effectively balance between leveraging calculated scores and exploring new samples, \texttt{OATS} adopts an \textbf{explore–exploit mechanism}. Specifically, the influence scores are stochastically re-evaluated to incorporate model training dynamics (\emph{``explore''}) while preserving previously identified high-quality data (\emph{``exploit''}). 

We summarize our contributions in this paper as follows:
\begin{itemize}[left=0pt, itemsep=0pt, topsep=0pt]
 \item \textbf{An Online Data Augmentation Paradigm for TSFM}. We propose a new paradigm for TSFM training that generates synthetic data tailored to each training step. This approach expands the potential of the traditional static data augmentation paradigm by taking the training process information into consideration.

 \item \textbf{A Novel Online Data Augmentation Strategy: \texttt{OATS}}. \texttt{OATS} leverages valuable training samples identified by data attribution scores as guiding signals and employs a diffusion model to generate synthetic data conditioned on these signals for each training step. Additionally, an explore-exploit mechanism is used to reduce computational cost and leverage local quality patterns. Overall, \texttt{OATS} offers a principled approach to dynamically generating high-quality synthetic data.
 
 \item \textbf{Comprehensive Empirical Evaluation}. We evaluate \texttt{OATS} on six evaluation datasets and two TSFM typical architectures (encoder-only and decoder-only TSFMs). \texttt{OATS} substantially outperforms the baseline data augmentation methods as well as the regular training.
\end{itemize}
\section{Method}\label{sec:method}

\begin{figure}[h]
    \centering
    \includegraphics[width=\linewidth]{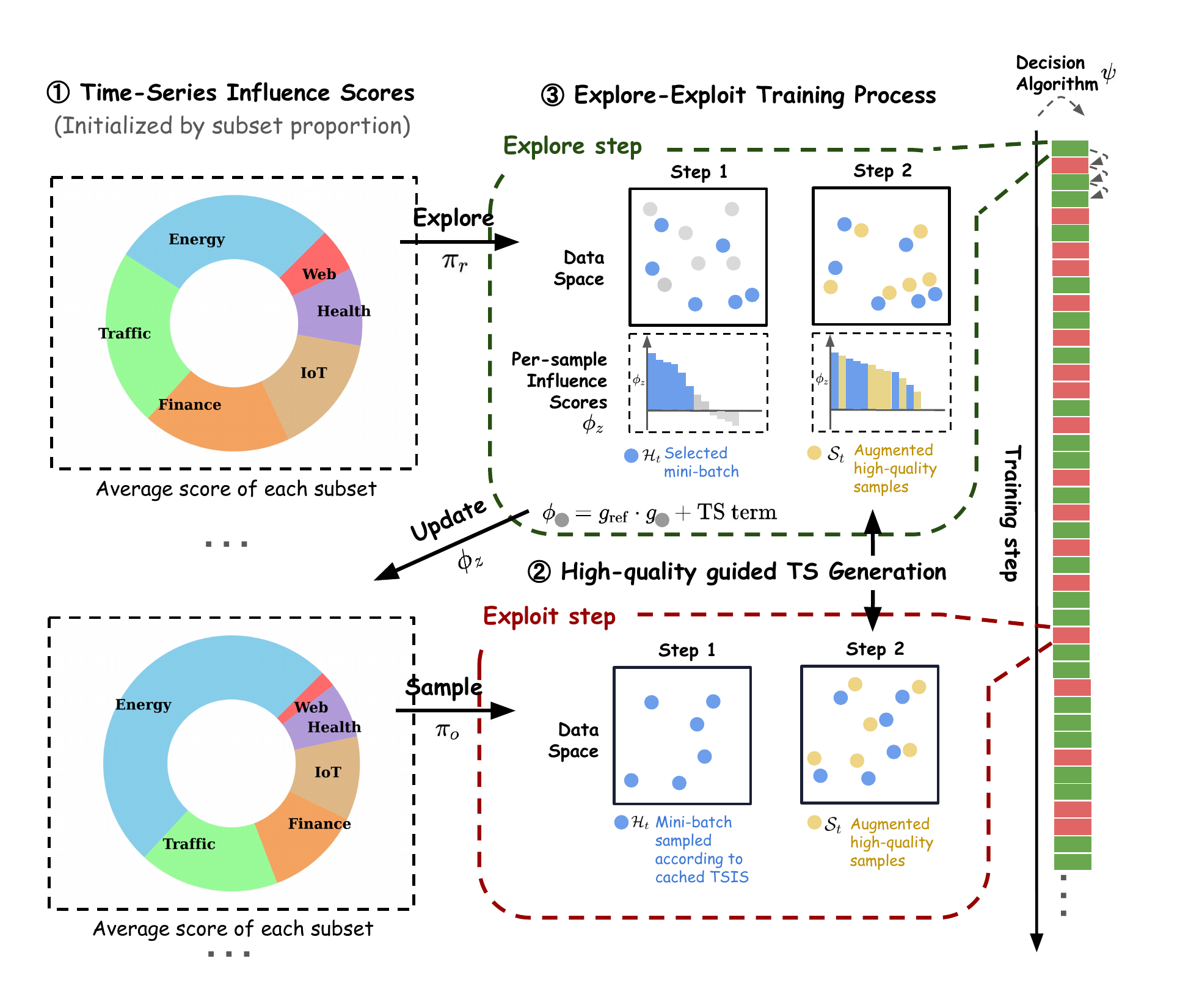}
    \caption{Architecture of \texttt{OATS}.  \texttt{OATS} employs three modules: \textcircled{1} Time-Series Influence Scores (TSIS) create generation guiding signals as high-quality data samples, and \textcircled{2} guide time series synthetic data generation for augmentation. \textcircled{3} Explore-exploit mechanism comprehensively plans if updating the TSIS or leveraging cached scores.}
    \label{fig:OATS}
\end{figure}

In this section, we introduce the modules of \underline{O}nline Data \underline{A}ugmentation for \underline{T}ime \underline{S}eries Foundation Models (\texttt{OATS}). \texttt{OATS} consists of three modules, i.e., \emph{Time-series Influence Scores (TSIS)} (Section~\ref{ssec:tsis}) quantitatively estimate the quality of time series samples and to identify valuable data as guiding signal; \emph{High-quality guided Data Generation} (Section~\ref{ssec:high-quality-guide-da}) generates synthetic samples conditioned on the guiding signal; \emph{Explore-exploit paradigm} (Section~\ref{ssec:explore-exploit}) balances the efficiency and effectiveness. We will introduce the modules in separate paragraphs. A diagram of the whole algorithm is presented in Figure~\ref{fig:OATS}, and an algorithm block is shown in Section~\ref{ssec:method-algorithm}.

\paragraph{Set-up of online data augmentation in TSFM.}

Suppose we have a training dataset $\mathcal{D}_{tr}$ of size $N$ which is partitioned into $L$ disjoint subsets: \(\mathcal{D}_{tr} = \bigcup_{l=1}^L\mathcal{D}_l = \{z_{l,k} | l=[L]; k=[N_l]\}\), where $N_l=|D_l|$ is the size of subset $\mathcal{D}_l$ and $z_{l,k} \in \mathcal{Z}$, the data space of time series samples. Subsets $\mathcal{D}_l$ are typically defined using domain metadata; for example, each subset represents a distinct environment, source, or distribution within the time series training data. We also have a reference dataset $\mathcal{D}_{ref}  = \{z_v | v=[N_{v}]\}$, $z_v \in \mathcal{Z}$. $\mathcal{D}_{ref}$ consists of a minimal number of samples (e.g., $N_v = 32$) and is strictly excluded from the model training process. It functions as a set of reference ``prompts'' to provide guiding signals for online augmentation.

A TSFM parameterized by $w \in \mathcal{W}$ is being trained on $\mathcal{D}_{tr}$ to minimize the loss function $\ell$ via an iterative optimization algorithm, e.g., stochastic gradient descent~\citep{ruder2016overview} for $T$ steps. The intermediate checkpoints of each step are represented as $\{w_t|t=[T]\}$. At iteration $t$, a mini-batch of data $\mathcal{B}_t = \{z_1, z_2, \ldots, z_B\}$ is sampled. Online data augmentation then generates synthetic samples $\mathcal{S}_t$, which are combined with all or part of $\mathcal{B}_t$ to update the model.

\subsection{Construct Guiding Signal via Time Series Influence Scores}\label{ssec:tsis}
Generating synthetic data is widely used in the training process, yet a core question remains unsettled: \emph{What should be generated?} A consistent and universally accepted answer has yet to emerge for the definition of the quality criteria for time series synthetic data in data augmentation. In \texttt{OATS}, we employ \emph{data attribution}~\citep{koh2017understanding,pruthi2020estimating} as a principled method to estimate the influence of a data point on the model output, which has shown significant usefulness in dataset optimization in different areas. The data attribution scores to reference loss are taken as the quantitative quality indicator.

We design a time-series influence score (TSIS)\footnote{We will use ``influence scores'' and ``data attribution scores'' interchangeably.} function with respect to a reference dataset $\mathcal{D}_{ref}$ as $\mathcal{F}_{\mathcal{D}_{ref}}$ for each sample to be
$\mathcal{F}_{\mathcal{D}_{ref}}:\mathcal{D}_{tr} \times \mathcal{W}\rightarrow \mathbb{R}$, which estimates the data attribution score of a data sample to be trained on $w_t$ with respect to the performance on $\mathcal{D}_{ref}$. A larger score indicates that training on $z$ for the next training step ($t+1$) leads to better performance on $\mathcal{D}_{ref}$. The TSIS function can be defined as:
\begin{align}
    \mathcal{F}_{\mathcal{D}_{ref}}(z_i, w_t) &= \underbrace{g_{\rm (ref);t} \cdot g_{z_i;t}}_{\rm Influence\ Score} - \underbrace{\mathbf{I}_{{\rm SNR}(z_i)<k}\cdot \infty}_{\text{TS-specific quality}}, \label{eq:itqs}
\end{align}
where \(g_{\rm (ref);t} = \nabla_{w} \sum_{v_i\in \mathcal{D}_{ref}}\ell(v_i, w_t) / |\mathcal{D}_{ref}|\) and \(g_{z_i;t} = \nabla_{w} \ell(z_i, w_t)\). The data attribution score is a first-order Taylor approximation of the reference loss change, where a careful derivation is stated in Proposition~\ref{pro:influence-score}. We additionally consider TS-specific quality indicators, which we use signal-to-noise ratio (SNR) in a pre-selection process, where $k$ is a threshold of minimal SNR and $\mathbf{I}$ is the indicator function. High-quality data $\mathcal{H}_t$ is identified as the samples with top-$k$ TSIS scores in each mini-batch $\mathcal{B}_t$ of training step $t$.

\begin{proposition}[First-order Taylor approximated influence score]\label{pro:influence-score}
The influence score term of TSIS is a first-order Taylor approximation of the difference of a utility function before and after a training step. Typically, we define the utility function to be the reference loss, which can be represented as
\begin{align*}
    \ell(\mathcal{D}_{ref}, w_t) &= \frac{1}{|\mathcal{D}_{ref}|}\sum_{v_i\in \mathcal{D}_{ref}}\ell(w_t, v_i)\\
    \ell(\mathcal{D}_{ref}, w_t) - \ell(\mathcal{D}_{ref}, w_{t+1}) 
    &= (w_t-w_{t+1})^\top \nabla_w\ell(\mathcal{D}_{ref}, w_t) + \mathcal{O}(\|w_{t+1}-w_t\|^2)\\
    &\simeq (w_t-w_{t+1})^\top \nabla_w\ell(\mathcal{D}_{ref}, w_t) \\
    &= \eta_t \nabla_w\ell(z_i, w_t)^\top \nabla_w\ell(\mathcal{D}_{ref}, w_t),
\end{align*}
where the model is updated through gradient descent\footnote{Here we use SGD on a single data sample, which has been widely accepted as a reasonable approximation for other optimizers like Adam in data attribution studies\citep{wang2024greats,wang2024capturing}.} on a data sample $z_i$ with learning rate $\eta_t$, i.e.,
\[
    w_{t+1} = w_t - \eta_t\nabla_w\ell(z_i, w_t).
\]
Given that the learning rate $\eta_t$ is typically small, the error of first-order Taylor approximation with level $\mathcal{O}(\|w_{t+1}-w_t\|^2)$ (or $\mathcal{O}(\eta^2)$ if the norm of gradient of loss is bounded) is small enough to provide an accurate estimate.
\end{proposition}

\begin{wrapfigure}{r}{0.55\textwidth}  
    \centering
    \includegraphics[width=0.5\textwidth]{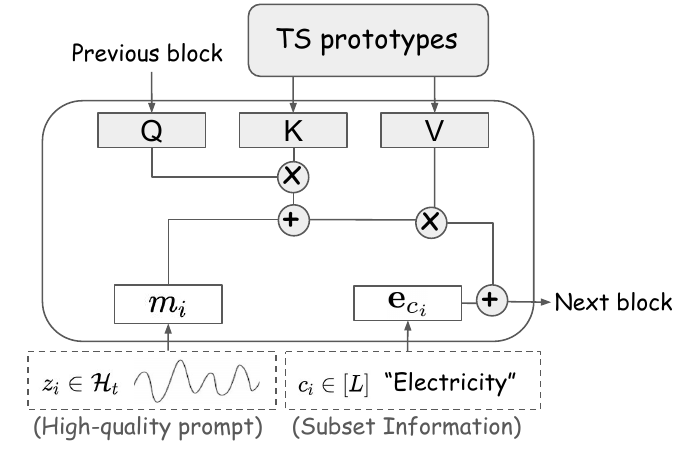}
    \caption{The architecture of the denoising diffusion model to generate synthetic data conditioned on constructed generation signals.}
    \label{fig:high-quality-diffusion}
\end{wrapfigure}

\subsection{High-quality Data Guided Data Generation}\label{ssec:high-quality-guide-da}

Relying solely on selected samples $\mathcal{H}_t$ limits diversity and risks overfitting. \texttt{OATS} uses the selected samples as guiding signals to generate realistic synthetic data, expanding the training data beyond simple oversampling. Once the guiding signal is constructed, a straightforward method is \emph{how can the synthetic data be generated?} We design a high-quality data guided generation model $\mathcal{G}$ for online synthetic data generation. We aim to model the conditional distribution \(p(\hat{z}|\mathcal{H}_t)\), which defines the probability of generating a sample \(\hat{z}\) given the guiding signal \(\mathcal{H}_{t} \subseteq \mathcal{B}_{t}\) for each training step $t$. The conditional generation model \(\mathcal{G}\) serves as a parameterized approximation of this distribution, enabling practical sampling via \(\hat{z}\sim \mathcal{G}(\mathcal{H}_{t})\). We design the architecture of $\mathcal{G}$ to be a diffusion model utilizing a time series semantic prototype module and take $\mathcal{H}_{t}$ as a generation condition. To incorporate the condition into the intermediate layers of the noise prediction network, we take \citet{huang2025timedp} as our backbone model.

We formulate the generation model $\mathcal{G}$ as a mixture model, where the time-series prototypes $P$ serve as the mixture components. By utilizing weights $\mathcal{M} = \{m_i|z_i\in \mathcal{H}_t\}$ extracted from the guiding signals $\mathcal{H}_t$, the diffusion model combines these prototypes to synthesize realistic data that captures diverse and representative temporal patterns. We additionally include a class conditional guidance based on the subset information to enhance the condition, denoted as $\mathcal{C}_{\rm high} = \{c_i| z_i\in \mathcal{H}_t\}$, where each $c_i \in [L]$ indicates which disjoint subset $\mathcal{D}_l$ the sample $z_i$ belongs to.

The generation (reverse diffusion) for each prompt can be expressed as
\begin{align}
    p(z^{(t-1)} | z^{(t)}, \mathcal{D}_{\rm high}) = \mathcal{N}(z^{(t-1)} | \mu(z^{(t)}, t, \underbrace{m_i, c_i}_{\mathclap{\rm High-quality\ data\ guidance}}), \sigma_t),
\end{align}
where $z^{(t)}$ is the intermediate sample at diffusion timestep $t$, $\mathcal{N}$ is Gaussian distribution, $\sigma_t$ is the noise covariance at diffusion step $t$. The architecture of the parameterized conditional noise predictor incorporates the prototype weights $m_i$ and class condition $c_i$. We present the architecture of the high-quality guided data augmentation in Figure~\ref{fig:high-quality-diffusion} in an intermediate $u^{\rm th}$ U-Net layer. The formula can be represented as
\begin{align}
    h^{(u)} = {\rm FF}({\rm Softmax}(\frac{QK^T}{\sqrt{d}}+ m_i)V || \mathbf{e}_{c_i}),
\end{align}
where \(Q=h^{(u-1)}W_Q^{(u)}\), \(K=PW_K^{(u)}\) and \(V=P W_V^{(u)}\) are the query, key, and value for the attention. $W_Q\in\mathbb{R}^{d\times d}$, $W_K\in\mathbb{R}^{d\times d}$, $W_V\in\mathbb{R}^{d\times d}$ are learnable parameters, $P\in \mathbb{R}^{N_p \times d}$ is the prototype latent arrays, $\mathbf{e}_{c_i} \in \mathbb{R}^{d_c}$ is the embedding of class $c_i$, $d$ and $d_c$ is the hidden dimension of prototype embedding and class embedding, and $FF$ is a feed-forward network. The final U-Net layer's output $h^{\rm final}$ is followed by another feed-forward network to produce the predicted noise.

\subsection{Explore \& Exploit Mechanism} \label{ssec:explore-exploit}

Ideally, \(\mathcal{H}_t\) should be identified from the whole training dataset $\mathcal{D}_{tr}$ in Section~\ref{ssec:tsis}, while the process of calculating TSIS for all training samples will be time-consuming. We are inspired by the multi-armed bandit problem and design an explore-exploit mechanism that could reuse the TSIS calculated for samples of previous training steps as well as explore new samples.

The mechanism divides training steps into \emph{explore} steps and \emph{exploit} steps. In \emph{explore} steps, we calculate TSIS for a batch of data samples $\mathcal{B}_t$ of size $b$ sampled by strategy $\pi_{explore}$ (hereafter denoted as $\pi_{r}$). We assume the locality of TSIS among $z$ in the same subset. Based on this assumption, we maintain a dynamic cache $\Phi_{\mathcal{D}_l}\in\mathbb{R},l\in[L]$ for each subset and update it through exponentially moving average. The partition of $\mathcal{D}_{tr} = \bigcup_{l=1}^L\mathcal{D}_l$ can naturally be the sub-datasets of a large collection or some clustering results. The update of $\Phi_{\mathcal{D}_l}$ on training step $t$ can be represented as
\begin{align}\label{eq:update-dataset-phi}
    \Phi_{\mathcal{D}_l} &= (1-\beta)\Phi_{\mathcal{D}_l} + \beta\sum_{z_i \in \mathcal{D}_l \cap \mathcal{B}_t} \mathcal{F}_{\mathcal{D}_{ref}}(z_i, w_t) / |\mathcal{D}_l \cap \mathcal{B}_t|,
\end{align}
where \(\mathcal{B}_t\) is a sample of size $b$ from \(\pi_r\), and \(\pi_r\) is a uniform sampling strategy \(\mathcal{U}(\mathcal{D}_{tr})\), $\beta \in [0,1]$ is the hyperparameter controls the decay factor of the exponentially moving average. The design of $\pi_r$ for exploring steps reflects the spirit to visit new data points and update the TS data. It randomly samples from the full training dataset $\mathcal{D}_{tr}$. Additionally, we refer to the influence score part of the TSIS in this subsection when we use $\mathcal{F}_{\mathcal{D}_{ref}}$.

For the \emph{exploit} step, we design a sample algorithm $\pi_{exploit}$ (hereafter denoted as $\pi_{o}$) utilizing the $\Phi_{\mathcal{D}_l}$ cached in \emph{explore} step. The design of $\pi_o$ reflects the latest estimation of each subset's quality.
\begin{align}
    \pi_o = \mathcal{U}(\mathcal{D}_l)\ \ \text{ with probability } \frac{|\mathcal{D}_l|\cdot\max(0,\Phi_{\mathcal{D}_l})}{\sum_{k}|\mathcal{D}_{k}|\cdot\max(0,\Phi_{\mathcal{D}_k})}
\end{align}

The choice of \emph{explore} or \emph{exploit} step is handled by $\epsilon$-greedy, a popular strategy for explore and exploit. We design a strategy $\psi$ that is defined to be ``explore'' with probability \(\epsilon\) and ``exploit'' with probability \((1-\epsilon)\), where $\epsilon$ is a hyperparameter controlling the balance between explore and exploit. Notably, the overhead of the \emph{exploit} step is small enough to be ignored since TSIS is not calculated in such steps. A more careful complexity analysis is introduced in Appendix~\ref{appendix:algorithm}.

\subsection{Algorithm.}\label{ssec:method-algorithm}

We summarize \underline{O}nline Data \underline{A}ugmentation for \underline{T}ime \underline{S}eries Foundation Models (\texttt{OATS}) and present in Algorithm~\ref{alg:oats}.

\begin{algorithm}
\caption{\underline{O}nline Data \underline{A}ugmentation for \underline{T}ime \underline{S}eries Foundation Models (\texttt{OATS})}\label{alg:oats}

\begin{algorithmic}[h]

\Require Training dataset and its $L$ disjoint subsets $\mathcal{D}_{tr} = \bigcup_{l=1}^L\mathcal{D}_l$, reference dataset $\mathcal{D}_{ref}$, training batch size $b$, conditional augmentation algorithm $\mathcal{G}$, training loss $\ell$, explore-exploit strategy $\psi$, explore sampling algorithm $\pi_{r}$, exploit sampling algorithm $\pi_{o}$, TSIS function $\mathcal{F}_{\mathcal{D}_{ref}}$.

\State Initialize model $w_0$
\State Initialize TSIS per subset $\Phi_{\mathcal{D}_l}, l \in [L]$ to be subset proportion.
\For{$t = 1$ to $T$}
    \If{$\psi(t)$ == ``explore''}
        \State Sample $\mathcal{B}_t \sim \pi_r^{b}$\Comment{Step 1}
    \State Calculate $\mathcal{F}_{\mathcal{D}_{ref}}(z_i, w_t)$ for $z_i\in\mathcal{B}_t$
     \State Update $\Phi_{\mathcal{D}_l}$ according to Equation~\ref{eq:update-dataset-phi}.
    \State Select a subset $\mathcal{H}_t \subseteq \mathcal{B}_t$ of samples with top-$\lfloor b/2 \rfloor$ of value $\mathcal{F}_{\mathcal{D}_{ref}}(z_i, w_t)$.
    \State Generate $\lfloor b/2 \rfloor$ samples as $\mathcal{S}_t$ using $\mathcal{G}$ guided by $\mathcal{H}_t$. \Comment{Step 2}
    \EndIf
    \If{$\psi(t)$ == ``exploit''}
        \State Sample $\mathcal{H}_t \sim \pi_o^{\lfloor b/2 \rfloor}$ \Comment{Step 1}
        \State Generate $\lfloor b/2 \rfloor$ samples as $\mathcal{S}_t$ using $\mathcal{G}$ guided by $\mathcal{H}_t$. \Comment{Step 2}
    \EndIf
    \State Update $w_t$ on mini-batch data $\mathcal{H}_t\ \cup\ \mathcal{S}_t$ and get $w_{t+1}$, continue to next step $t+1$.
\EndFor
\end{algorithmic}
\label{alg:oats}
\end{algorithm}
\section{Experiments}\label{sec:experiments}

In this section, we present the empirical evaluation of \texttt{OATS}. We first
introduce the experiment setup in Section~\ref{ssec:experiment-setup}. We then evaluate the performance of \texttt{OATS} on typical TSFM architectures and datasets in Section~\ref{ssec:performance-oats}. In addition, we examine the performance of \texttt{OATS} in different explore-exploit ratio in Section~\ref{ssec:epsilon} and show some case studies in Section~\ref{ssec:case-studies}.

\subsection{Experiment Setup}\label{ssec:experiment-setup}

\vpara{Model.} We conduct experiments on two typical TSFM architectures, i.e., encoder-only and decoder-only Transformer. We follow the same training settings and model definitions, which incorporate patch embedding, rotary positional embedding, and a mixture of distributions to better adapt to time series forecasting while preserving extensibility, in \citet{yao2024towards} to match the popular TSFM forecasting models, Moirai~\citep{woo2024unified} and Chronos~\citep{ansari2024chronos}. We will call these two models ``Encoder-only TSFM'' and ``Decoder-only TSFM'' in our following experiments.

\vpara{Datasets.} Following \citet{yao2024towards}, we train ``Encoder-only TSFM'' and ``Decoder-only TSFM'' on the LOTSA dataset in the pretraining stage. These models are then evaluated on the LSF dataset~\citep{wu2023timesnet}, using various prediction lengths and a preprocessing pipeline as in \citet{yao2024towards}. We evaluate 6 datasets in LSF~\citep{wu2023timesnet} (ETTm1, ETTm2, ETTh1, ETTh2, Weather, Electricity). We also take a very small number of samples (32 in all our experiments) from these evaluation datasets as a reference set used by TSIS. The training dataset of the high-quality prompt-guided diffusion model is a very small subset sampled from the training dataset of TSFM. In our experiments, we sample 5\% of the training data.

\vpara{Baselines.} We examine two popular data augmentation methods for TSFM. ``TSMixup'' or Time Series Mixup is proposed in \citet{ansari2024chronos} which creates new data samples from $k$ existing ones through weighted summation. The generation process can be represented as \(\hat{z} = \sum_{i=1}^k\lambda_iz_i\), where $\hat{z}$ is the generated sample and $\lambda_i$ is the weight for the $i^{\rm th}$ sample. ``Jitter''~\citep{um2017data} is another widely adopted data augmentation method that involves small noise on existing samples to increase the robustness and diversity. The generation process can be represented as \(\hat{z} = z + \epsilon\), where $\epsilon\sim\mathcal{N}(0,\sigma)$ and $\sigma$ is the noise variance. We also include the result of the regular training process without additional data augmentation.

\vpara{Evaluation Metrics.} To be consistent with \citet{yao2024towards}, we primarily report the normalized mean absolute percentage error (MAPE) and negative log-likelihood (NLL) for Encoder-only TSFM and Decoder-only TSFM, as these metrics avoid distortions caused by high-amplitude samples. We specifically report the metrics for prediction length to be 192 and ``overall'' (average performance of using prediction lengths). Here we highlight 192 because \cite{yao2024towards} takes 192 as prediction length.

\subsection{Performance of \texttt{OATS}}\label{ssec:performance-oats}

\begin{figure}[htbp]
    \centering
    \begin{subfigure}[b]{0.24\textwidth}
        \centering
        \includegraphics[width=\textwidth]{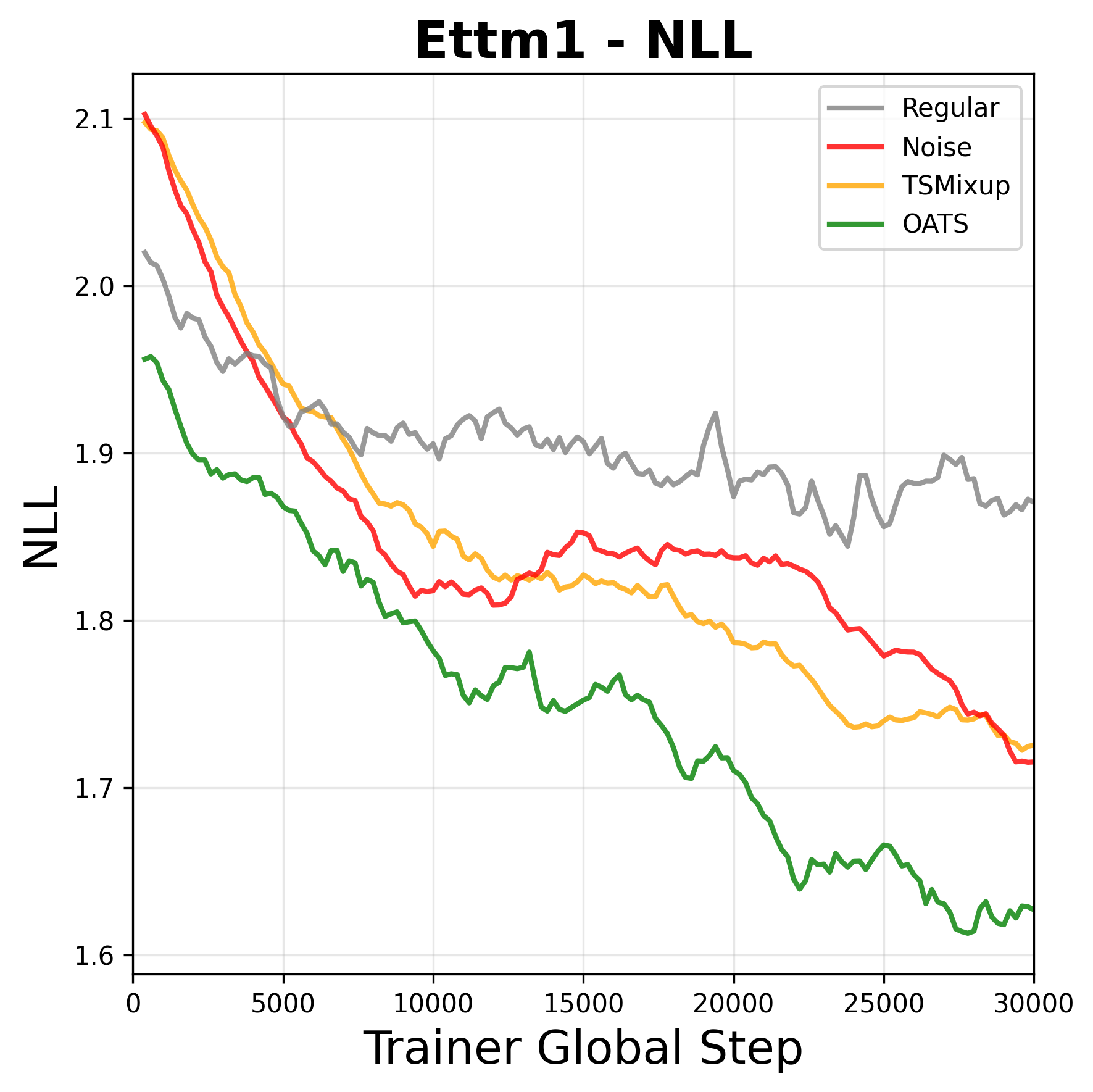}
        \caption{ETTm1}
        \label{fig:3}
    \end{subfigure}
    \begin{subfigure}[b]{0.24\textwidth}
        \centering
        \includegraphics[width=\textwidth]{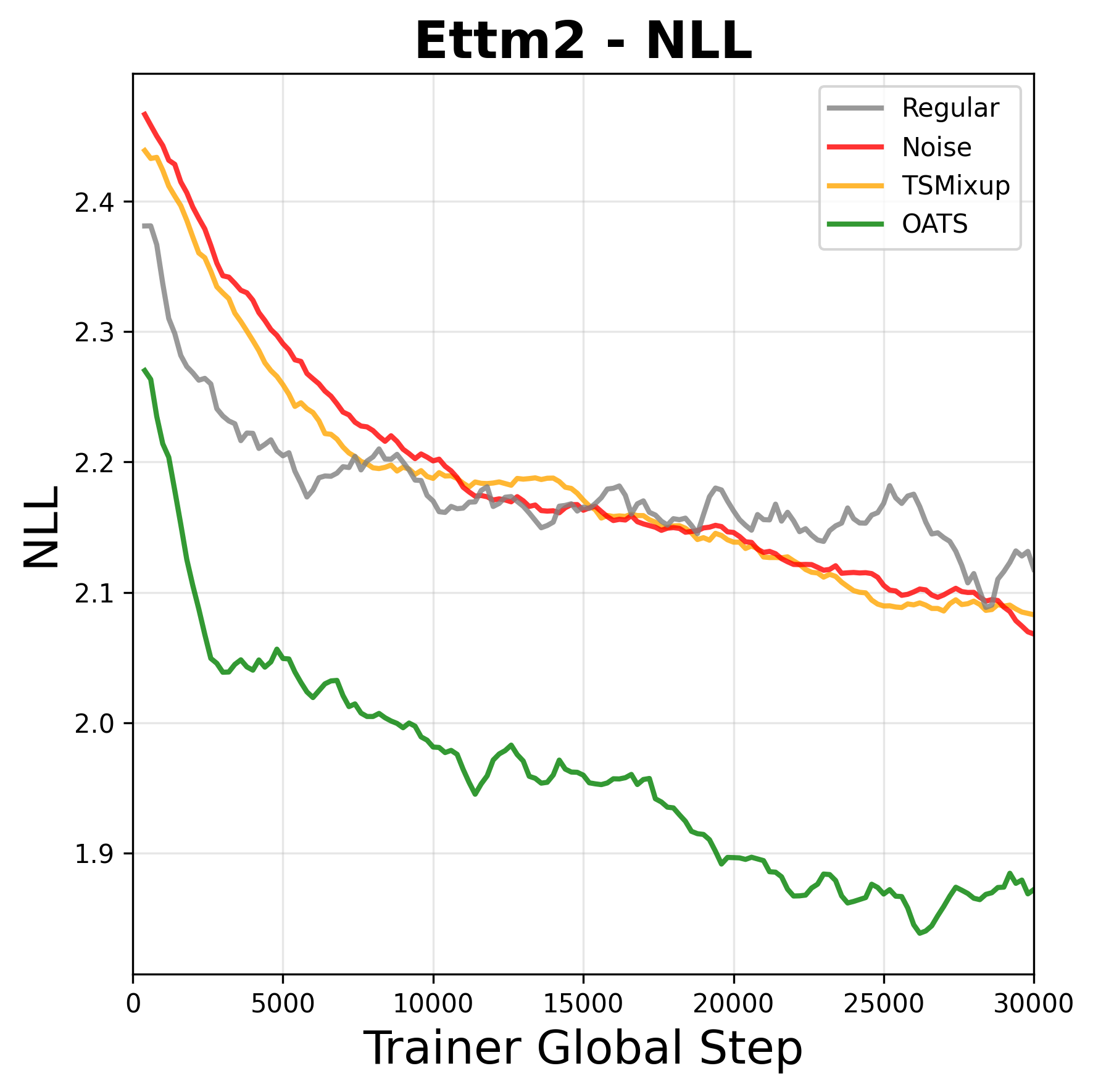}
        \caption{ETTm2}
        \label{fig:4}
    \end{subfigure}
    \begin{subfigure}[b]{0.24\textwidth}
        \centering
        \includegraphics[width=\textwidth]{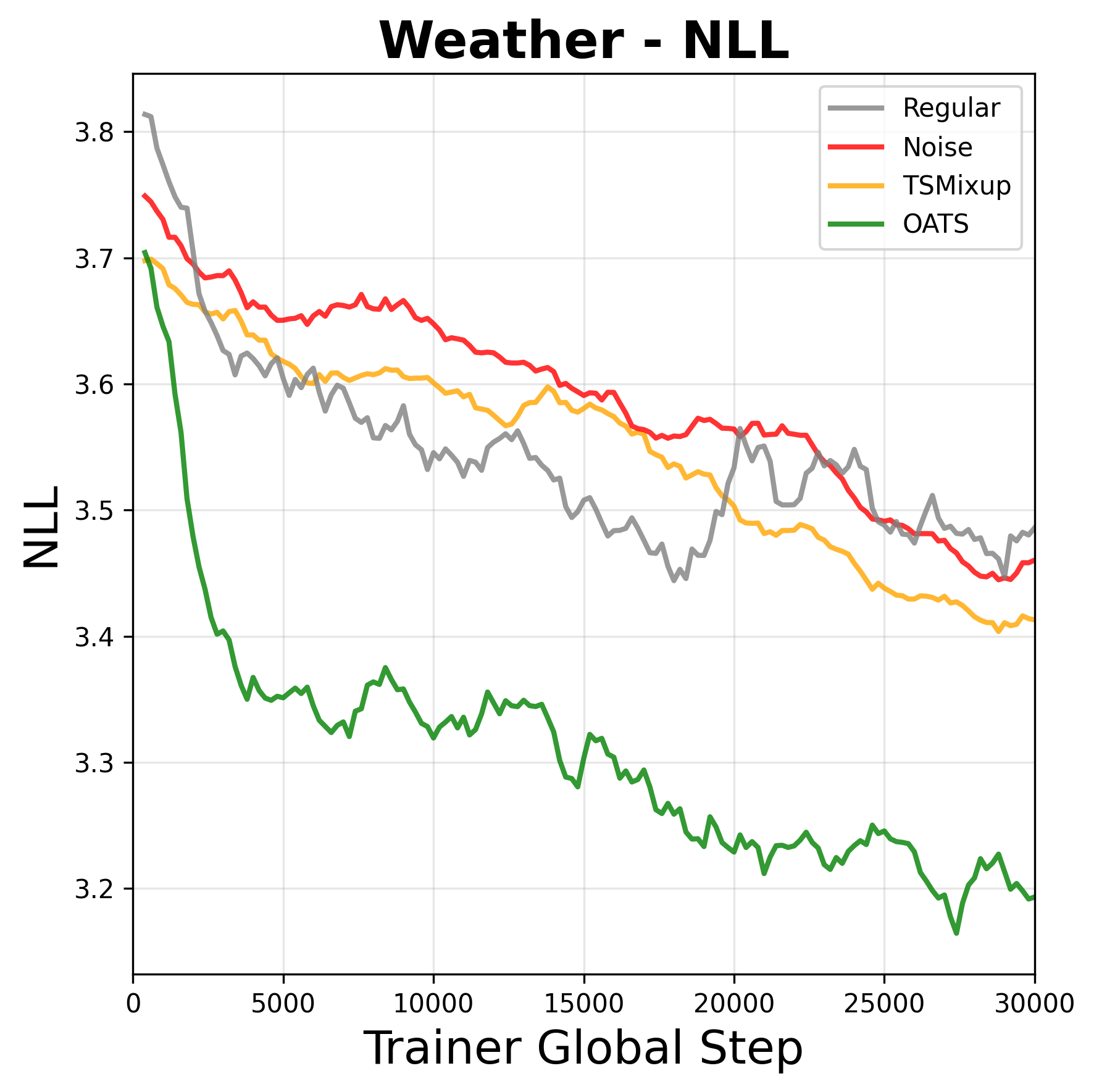}
        \caption{Weather}
        \label{fig:5}
    \end{subfigure}
    \begin{subfigure}[b]{0.24\textwidth}
        \centering
        \includegraphics[width=\textwidth]{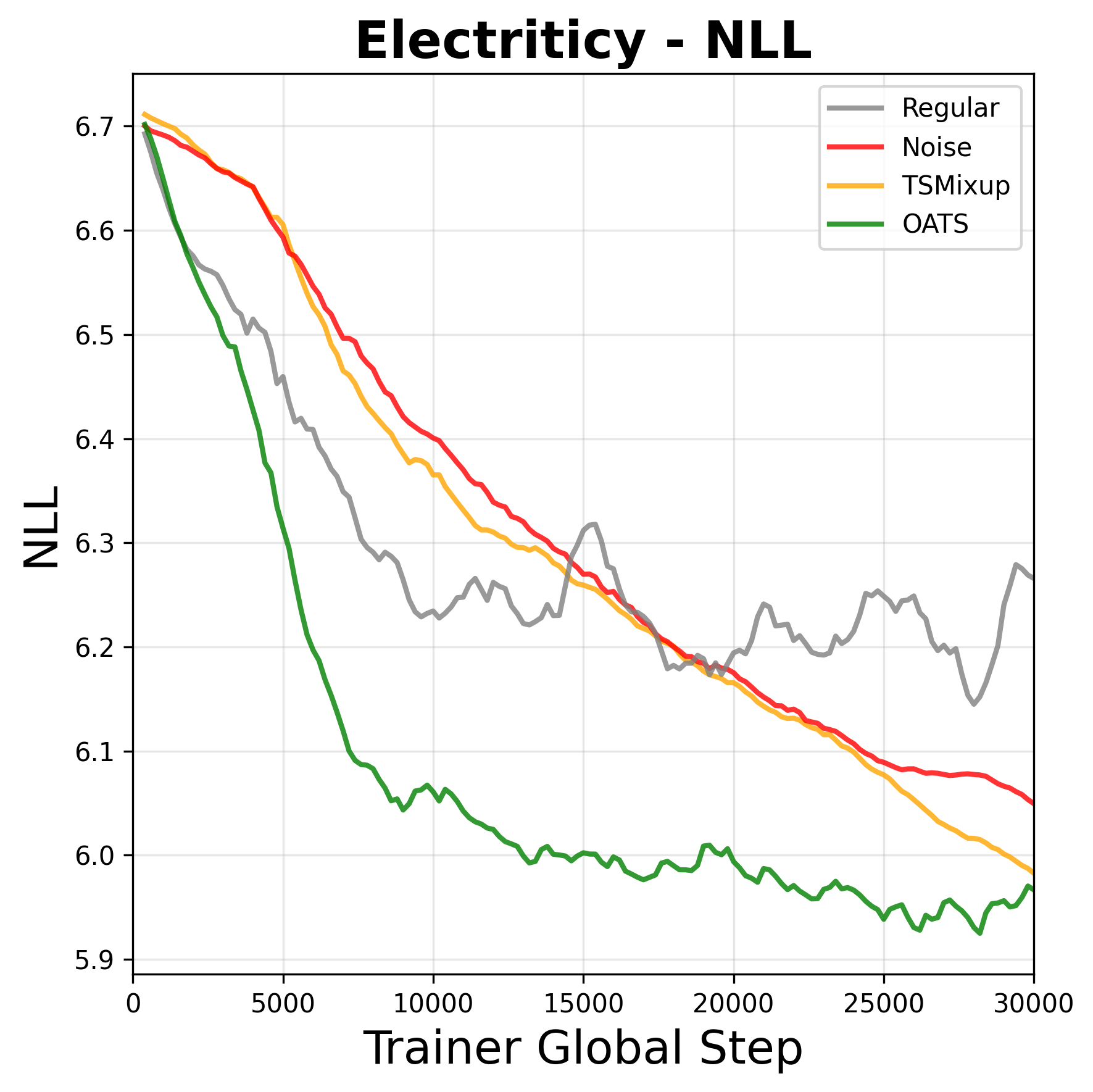}
        \caption{Electricity}
        \label{fig:6}
    \end{subfigure}
    \caption{Test loss (NLL) of \texttt{OATS}, TSMixup, Jitter and Regular training for each training step.}\label{fig:training-curve-test-loss}
\end{figure}

In Table~\ref{table:encoder_performance_overall} and Table~\ref{table:decoder_performance_overall}, we present the NLL and MAPE on the test datasets on Encoder-only TSFM and Decoder-only TSFM, respectively. \texttt{OATS} outperforms baselines as well as regular training in almost all cases and both metrics. Especially, for datasets ETTm1, ETTm2, ETTh2, Weather and Electricity, \texttt{OATS} reaches the best result. TSMixup and Jitter get mixed performance compared to regular training, as presented by the light green and red backgrounds. While \texttt{OATS} performs more consistently better than regular training. In Figure~\ref{fig:training-curve-test-loss}, we present the test loss (NLL) curve on evaluation datasets on Encoder-only TSFM and different test datasets. \texttt{OATS} achieves a faster reduction in NLL than other baseline data augmentation methods, as well as the regular training, and achieves better overall performance. Additional results are provided in Appendix~\ref{app:additional-results}.

\begin{table}[h]
\caption{Performance comparison of various data augmentation methods on encoder-only TSFM with prediction length be 192 / average over all prediction lengths) and $\epsilon=1$. \textbf{Bold} means the best result. \colorbox{green!15}{Light green} background means that the performance is better than the regular training process, while \colorbox{red!10}{light red} background means that the performance is worse than the regular training process. The error bar shows the standard error of the mean over 5 independent runs.}\label{table:encoder_performance_overall}
\resizebox{\textwidth}{!}{
\begin{tabular}{@{}l|l|llllllll@{}}
\toprule
\multirow{3}{*}{Dataset} & \multirow{3}{*}{Pred. length} & \multicolumn{2}{l}{\multirow{2}{*}{OATS}} & \multicolumn{2}{l}{\multirow{2}{*}{TSMixup}} & \multicolumn{2}{l}{\multirow{2}{*}{Jitter}} & \multicolumn{2}{l}{\multirow{2}{*}{Regular}} \\
                         &                               & \multicolumn{2}{l}{}                        & \multicolumn{2}{l}{}                    & \multicolumn{2}{l}{}                   & \multicolumn{2}{l}{}                       \\ \cmidrule(lr){3-10}
                         &                               & NLL                & MAPE                & NLL          & MAPE         & NLL          & MAPE         & NLL               & MAPE               \\ \midrule
\multirow{2}{*}{ETTm1} & 192 & \cellcolor{green!15}\textbf{1.627 $\pm$ 0.042} & \cellcolor{green!15}\textbf{0.672 $\pm$ 0.043} & \cellcolor{green!15}1.725 $\pm$ 0.031 & \cellcolor{green!15}0.759 $\pm$ 0.038 & \cellcolor{green!15}1.715 $\pm$ 0.047 & \cellcolor{green!15}0.691 $\pm$ 0.044 & 1.870 $\pm$ 0.019 & 0.844 $\pm$ 0.056 \\
 & Overall & \cellcolor{green!15}\textbf{1.614 $\pm$ 0.020} & \cellcolor{green!15}\textbf{0.623 $\pm$ 0.024} & \cellcolor{green!15}1.715 $\pm$ 0.018 & \cellcolor{green!15}0.700 $\pm$ 0.019 & \cellcolor{green!15}1.731 $\pm$ 0.025 & \cellcolor{green!15}0.650 $\pm$ 0.019 & 1.854 $\pm$ 0.016 & 0.783 $\pm$ 0.037 \\
\midrule
\multirow{2}{*}{ETTm2} & 192 & \cellcolor{green!15}\textbf{1.872 $\pm$ 0.014} & \cellcolor{green!15}\textbf{0.208 $\pm$ 0.005} & \cellcolor{green!15}2.083 $\pm$ 0.016 & \cellcolor{red!10}0.271 $\pm$ 0.006 & \cellcolor{green!15}2.068 $\pm$ 0.020 & \cellcolor{green!15}0.256 $\pm$ 0.007 & 2.118 $\pm$ 0.028 & 0.267 $\pm$ 0.009 \\
 & Overall & \cellcolor{green!15}\textbf{1.863 $\pm$ 0.013} & \cellcolor{green!15}\textbf{0.222 $\pm$ 0.004} & \cellcolor{green!15}2.097 $\pm$ 0.011 & \cellcolor{red!10}0.292 $\pm$ 0.003 & \cellcolor{green!15}2.072 $\pm$ 0.009 & \cellcolor{green!15}0.273 $\pm$ 0.005 & 2.128 $\pm$ 0.014 & 0.290 $\pm$ 0.008 \\
\midrule
\multirow{2}{*}{ETTh1} & 192 & \cellcolor{green!15}\textbf{1.794 $\pm$ 0.035} & \cellcolor{green!15}0.681 $\pm$ 0.025 & \cellcolor{green!15}1.883 $\pm$ 0.030 & \cellcolor{green!15}0.634 $\pm$ 0.025 & \cellcolor{red!10}1.959 $\pm$ 0.028 & \cellcolor{green!15}\textbf{0.629 $\pm$ 0.035} & 1.897 $\pm$ 0.040 & 0.758 $\pm$ 0.059 \\
 & Overall & \cellcolor{green!15}\textbf{1.814 $\pm$ 0.020} & \cellcolor{green!15}0.709 $\pm$ 0.021 & \cellcolor{green!15}1.860 $\pm$ 0.027 & \cellcolor{green!15}\textbf{0.634 $\pm$ 0.011} & \cellcolor{green!15}1.913 $\pm$ 0.026 & \cellcolor{green!15}0.644 $\pm$ 0.021 & 1.915 $\pm$ 0.019 & 0.812 $\pm$ 0.031 \\
\midrule
\multirow{2}{*}{ETTh2} & 192 & \cellcolor{green!15}\textbf{1.857 $\pm$ 0.063} & \cellcolor{green!15}\textbf{0.267 $\pm$ 0.007} & \cellcolor{red!10}2.123 $\pm$ 0.029 & \cellcolor{red!10}0.294 $\pm$ 0.011 & \cellcolor{red!10}2.127 $\pm$ 0.022 & \cellcolor{red!10}0.296 $\pm$ 0.013 & 2.084 $\pm$ 0.047 & 0.282 $\pm$ 0.015 \\
 & Overall & \cellcolor{green!15}\textbf{1.876 $\pm$ 0.025} & \cellcolor{green!15}\textbf{0.263 $\pm$ 0.005} & \cellcolor{red!10}2.113 $\pm$ 0.020 & \cellcolor{red!10}0.294 $\pm$ 0.005 & \cellcolor{red!10}2.122 $\pm$ 0.022 & \cellcolor{red!10}0.298 $\pm$ 0.005 & 2.086 $\pm$ 0.021 & 0.283 $\pm$ 0.008 \\
\midrule
\multirow{2}{*}{Weather} & 192 & \cellcolor{green!15}\textbf{3.193 $\pm$ 0.043} & \cellcolor{green!15}\textbf{1.778 $\pm$ 0.127} & \cellcolor{green!15}3.413 $\pm$ 0.041 & \cellcolor{green!15}2.016 $\pm$ 0.040 & \cellcolor{green!15}3.460 $\pm$ 0.043 & \cellcolor{red!10}2.619 $\pm$ 0.280 & 3.486 $\pm$ 0.051 & 2.331 $\pm$ 0.285 \\
 & Overall & \cellcolor{green!15}\textbf{3.216 $\pm$ 0.036} & \cellcolor{green!15}\textbf{1.659 $\pm$ 0.069} & \cellcolor{green!15}3.428 $\pm$ 0.023 & \cellcolor{green!15}1.948 $\pm$ 0.033 & \cellcolor{green!15}3.499 $\pm$ 0.020 & \cellcolor{red!10}2.591 $\pm$ 0.145 & 3.526 $\pm$ 0.026 & 2.267 $\pm$ 0.125 \\
\midrule
\multirow{2}{*}{Electricity} & 192 & \cellcolor{green!15}\textbf{5.967 $\pm$ 0.021} & \cellcolor{green!15}\textbf{0.515 $\pm$ 0.071} & \cellcolor{green!15}5.983 $\pm$ 0.010 & \cellcolor{green!15}0.553 $\pm$ 0.034 & \cellcolor{green!15}6.050 $\pm$ 0.015 & \cellcolor{green!15}0.659 $\pm$ 0.062 & 6.266 $\pm$ 0.019 & 0.750 $\pm$ 0.033 \\
 & Overall & \cellcolor{green!15}\textbf{5.945 $\pm$ 0.011} & \cellcolor{green!15}\textbf{0.507 $\pm$ 0.045} & \cellcolor{green!15}5.987 $\pm$ 0.007 & \cellcolor{green!15}0.555 $\pm$ 0.013 & \cellcolor{green!15}6.044 $\pm$ 0.010 & \cellcolor{green!15}0.642 $\pm$ 0.025 & 6.260 $\pm$ 0.015 & 0.767 $\pm$ 0.021 \\
\bottomrule
\end{tabular}
}
\end{table}

\begin{table}[h]
\caption{Performance comparison of various data augmentation methods on decoder-only TSFM with prediction length being 192 / average over all prediction lengths) and $\epsilon=1$. \textbf{Bold} means the best result. \colorbox{green!15}{Light green} background means that the performance is better than the regular training process, while \colorbox{red!10}{light red} background means that the performance is worse than the regular training process.}\label{table:decoder_performance_overall}
\resizebox{\textwidth}{!}{
\begin{tabular}{@{}l|l|llllllll@{}}
\toprule
\multirow{3}{*}{Dataset} & \multirow{3}{*}{Pred. length} & \multicolumn{2}{l}{\multirow{2}{*}{OATS}} & \multicolumn{2}{l}{\multirow{2}{*}{TSMixup}} & \multicolumn{2}{l}{\multirow{2}{*}{Jitter}} & \multicolumn{2}{l}{\multirow{2}{*}{Regular}} \\
                         &                               & \multicolumn{2}{l}{}                        & \multicolumn{2}{l}{}                    & \multicolumn{2}{l}{}                   & \multicolumn{2}{l}{}                       \\ \cmidrule(lr){3-10}
                         &                               & NLL                & MAPE                & NLL          & MAPE         & NLL          & MAPE         & NLL               & MAPE               \\ \midrule
\multirow{2}{*}{ETTm1} & 192 & \cellcolor{green!15}\textbf{1.654 $\pm$ 0.018} & \cellcolor{green!15}\textbf{0.659 $\pm$ 0.027} & \cellcolor{green!15}1.691 $\pm$ 0.013 & \cellcolor{green!15}0.664 $\pm$ 0.027 & \cellcolor{red!10}1.767 $\pm$ 0.041 & \cellcolor{green!15}0.675 $\pm$ 0.031 & 1.737 $\pm$ 0.034 & 0.677 $\pm$ 0.022 \\
 & Overall & \cellcolor{green!15}\textbf{1.740 $\pm$ 0.015} & \cellcolor{green!15}\textbf{0.635 $\pm$ 0.014} & \cellcolor{green!15}1.785 $\pm$ 0.025 & \cellcolor{green!15}0.649 $\pm$ 0.016 & \cellcolor{red!10}1.842 $\pm$ 0.034 & \cellcolor{green!15}0.650 $\pm$ 0.018 & 1.812 $\pm$ 0.022 & 0.653 $\pm$ 0.015 \\
\midrule
\multirow{2}{*}{ETTm2} & 192 & \cellcolor{green!15}\textbf{1.726 $\pm$ 0.039} & \cellcolor{green!15}\textbf{0.202 $\pm$ 0.003} & \cellcolor{green!15}1.789 $\pm$ 0.014 & \cellcolor{green!15}0.206 $\pm$ 0.002 & \cellcolor{red!10}1.857 $\pm$ 0.044 & \cellcolor{red!10}0.222 $\pm$ 0.007 & 1.841 $\pm$ 0.009 & 0.217 $\pm$ 0.001 \\
 & Overall & \cellcolor{green!15}\textbf{1.765 $\pm$ 0.030} & \cellcolor{green!15}\textbf{0.228 $\pm$ 0.002} & \cellcolor{green!15}1.824 $\pm$ 0.015 & \cellcolor{green!15}0.229 $\pm$ 0.002 & \cellcolor{red!10}1.880 $\pm$ 0.034 & \cellcolor{red!10}0.244 $\pm$ 0.008 & 1.874 $\pm$ 0.008 & 0.242 $\pm$ 0.004 \\
\midrule
\multirow{2}{*}{ETTh1} & 192 & \cellcolor{green!15}\textbf{1.759 $\pm$ 0.046} & \cellcolor{green!15}0.534  $\pm$ 0.022 & \cellcolor{red!10}1.933 $\pm$ 0.069 & \cellcolor{green!15}0.536 $\pm$ 0.017 & \cellcolor{green!15}1.808 $\pm$ 0.069 & \cellcolor{green!15}\textbf{0.504 $\pm$ 0.021} & 1.852 $\pm$ 0.046 & 0.537 $\pm$ 0.026 \\
 & Overall & \cellcolor{green!15}\textbf{1.824 $\pm$ 0.020} & \cellcolor{green!15}0.562  $\pm$ 0.022 & \cellcolor{red!10}2.038 $\pm$ 0.031 & \cellcolor{red!10}0.568 $\pm$ 0.014 & \cellcolor{green!15}1.878 $\pm$ 0.030 & \cellcolor{green!15}\textbf{0.538 $\pm$ 0.016} & 1.943 $\pm$ 0.025 & 0.563 $\pm$ 0.013 \\
\midrule
\multirow{2}{*}{ETTh2} & 192 & \cellcolor{green!15}\textbf{1.817 $\pm$ 0.045} & \cellcolor{green!15}\textbf{0.261 $\pm$ 0.011} & \cellcolor{red!10}1.977 $\pm$ 0.036 & \cellcolor{red!10}0.289 $\pm$ 0.009 & \cellcolor{red!10}1.923 $\pm$ 0.036 & \cellcolor{green!15}0.275 $\pm$ 0.014 & 1.890 $\pm$ 0.050 & 0.284 $\pm$ 0.008 \\
 & Overall & \cellcolor{green!15}\textbf{1.936 $\pm$ 0.026} & \cellcolor{green!15}\textbf{0.274 $\pm$ 0.007} & \cellcolor{red!10}2.087 $\pm$ 0.030 & \cellcolor{green!15}0.307 $\pm$ 0.005 & \cellcolor{green!15}1.994 $\pm$ 0.022 & \cellcolor{green!15}0.295 $\pm$ 0.008 & 2.027 $\pm$ 0.023 & 0.309 $\pm$ 0.006 \\
\midrule
\multirow{2}{*}{Weather} & 192 & \cellcolor{green!15}\textbf{2.914 $\pm$ 0.081} & \cellcolor{green!15}\textbf{2.885 $\pm$ 0.650} & \cellcolor{red!10}2.974 $\pm$ 0.059 & \cellcolor{red!10}3.935 $\pm$ 0.334 & \cellcolor{red!10}3.025 $\pm$ 0.065 & \cellcolor{red!10}5.423 $\pm$ 0.822 & 2.914 $\pm$ 0.026 & 3.059 $\pm$ 0.622 \\
 & Overall & \cellcolor{green!15}\textbf{3.168 $\pm$ 0.058} & \cellcolor{green!15}\textbf{2.376 $\pm$ 0.323} & \cellcolor{red!10}3.265 $\pm$ 0.039 & \cellcolor{red!10}3.643 $\pm$ 0.186 & \cellcolor{red!10}3.276 $\pm$ 0.037 & \cellcolor{red!10}4.624 $\pm$ 0.411 & 3.245 $\pm$ 0.041 & 2.946 $\pm$ 0.574 \\
\midrule
\multirow{2}{*}{Electricity} & 192 & \cellcolor{green!15}\textbf{6.041 $\pm$ 0.017} & \cellcolor{green!15}\textbf{0.528 $\pm$ 0.030} & \cellcolor{red!10}6.074 $\pm$ 0.033 & \cellcolor{green!15}0.563 $\pm$ 0.010 & \cellcolor{red!10}6.090 $\pm$ 0.022 & \cellcolor{red!10}0.627 $\pm$ 0.028 & 6.049 $\pm$ 0.026 & 0.588 $\pm$ 0.020 \\
 & Overall & \cellcolor{green!15}\textbf{6.040 $\pm$ 0.011} & \cellcolor{green!15}\textbf{0.526 $\pm$ 0.014} & \cellcolor{red!10}6.079 $\pm$ 0.018 & \cellcolor{green!15}0.564 $\pm$ 0.014 & \cellcolor{red!10}6.107 $\pm$ 0.019 & \cellcolor{red!10}0.631 $\pm$ 0.018 & 6.057 $\pm$ 0.015 & 0.579 $\pm$ 0.011 \\
\bottomrule
\end{tabular}
}
\end{table}

\subsection{Performance of \texttt{OATS} under Different Explore-Exploit Levels}\label{ssec:epsilon}

The key experiment goal of this subsection is to examine the sensitivity and behavior of \texttt{OATS} on different levels of explore-exploit mechanisms through adjusting the value of $\epsilon$. $\epsilon$ is a hyperparameter designed to control the possibility of carrying out \emph{explore} ($\epsilon$) or \emph{exploit} step ($1-\epsilon$).  \emph{Explore} step will examine new data by calculating the TSIS ($\mathcal{F}_{\mathcal{D}_{ref}}$), which is time-consuming. \emph{Exploit} step will leverage cached TSIS ($\Phi_{\mathcal{D}_l}$) calculated in previous steps to directly sample a high-quality batch, which is light-weighted. 

\begin{figure}[h]
    \vspace{-5pt}
    \centering
    \includegraphics[width=\linewidth]{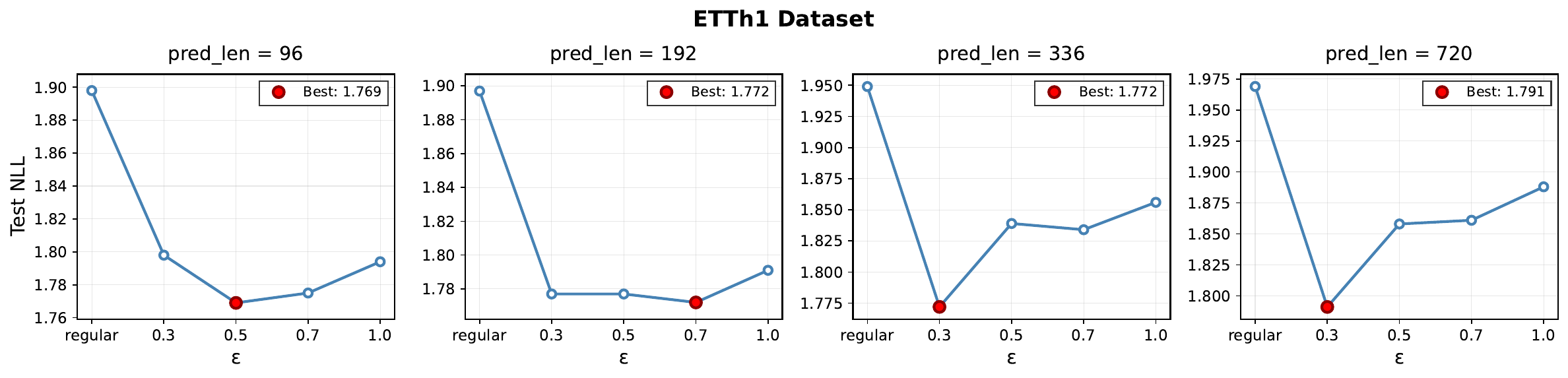}
    \includegraphics[width=\linewidth]{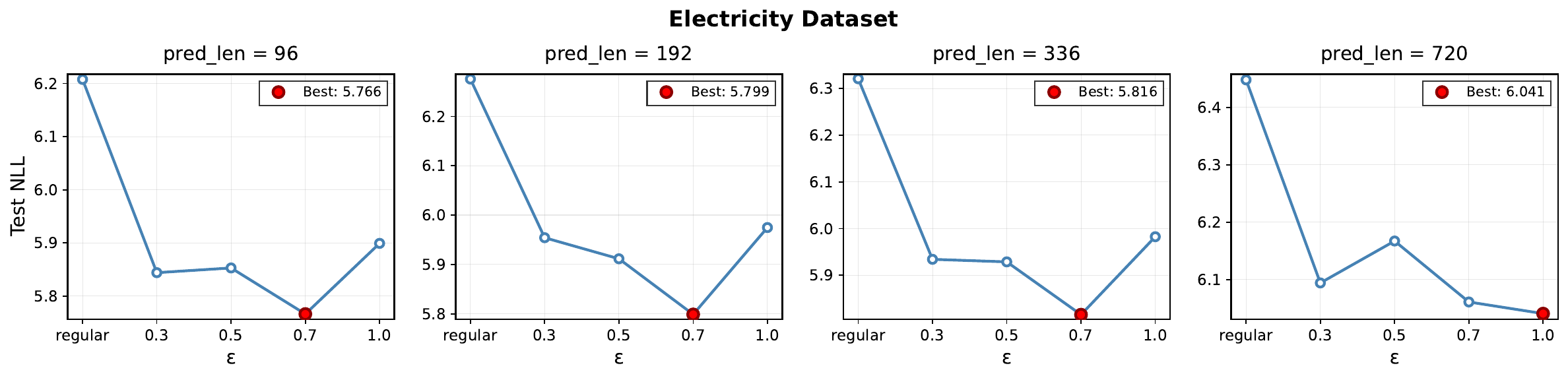}
    \caption{Performance of \texttt{OATS} on different explore-exploit ratio $\epsilon$. (\textbf{First Row}) Test NLL on ETTh1. (\textbf{Second Row}) Test NLL on Electricity. The best performance is labeled by a red circle.}
    \label{fig:epsilon}
\end{figure}

In Figure~\ref{fig:epsilon} we present the performance of \texttt{OATS} on different levels of explore-exploit mechanisms. We choose to set $\epsilon = [0.3, 0.5, 0.7, 1.0]$ and report the performance on two test datasets (ETTh1 and Electricity) and four prediction lengths (96, 192, 336, 720).

The best performance in each setting is often not achieved by $\epsilon=1$ (all explore step), the most computationally heavy setting that calculates TSIS in every step. This shows that leveraging cached TSIS is helpful to select higher-quality batch (using $\pi_o$) to guide the data augmentation and demonstrates the potential of using the mechanism to reduce computational overhead. Another observation also shows that \texttt{OATS} consistently performs better than the regular training process regardless of the setting of $\epsilon$. We included a detailed complexity and time cost analysis in Appendix~\ref{appendix:algorithm}.

\subsection{Case studies}\label{ssec:case-studies}

In this subsection, we carry out two case studies to further provide intuitive evidence of our motivation and some insights.

\begin{figure}[h]
    \centering
    \begin{subfigure}{0.48\textwidth}
        \centering
        \includegraphics[width=\linewidth]{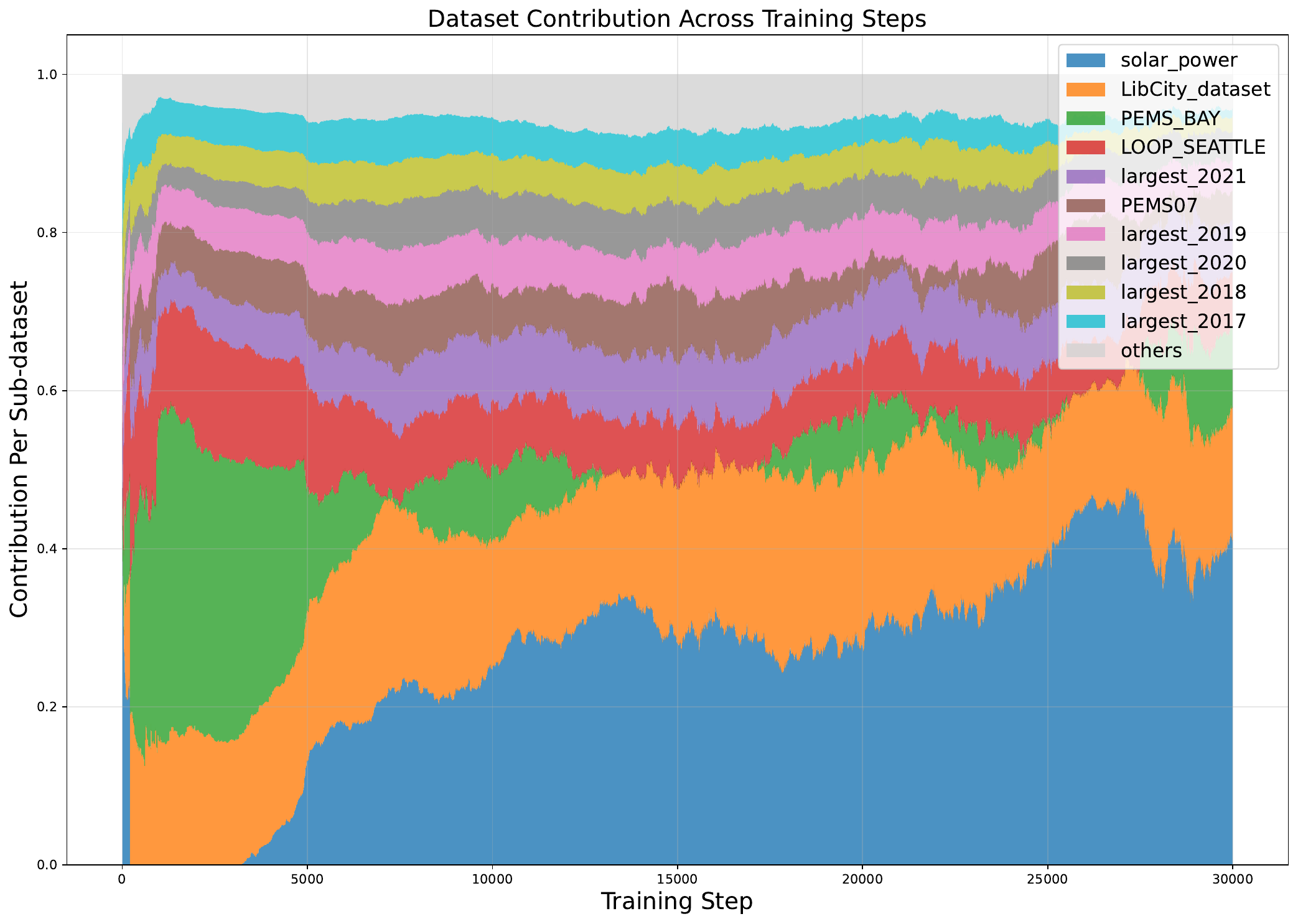} 
        \caption{Sub-dataset TSIS ($\Phi_{\mathcal{D}_l}$) along the training.}
        \label{fig:case1-per-subset-TSIS}
    \end{subfigure}
    \hfill
    \begin{subfigure}{0.48\textwidth}
        \centering
        \includegraphics[width=\linewidth]{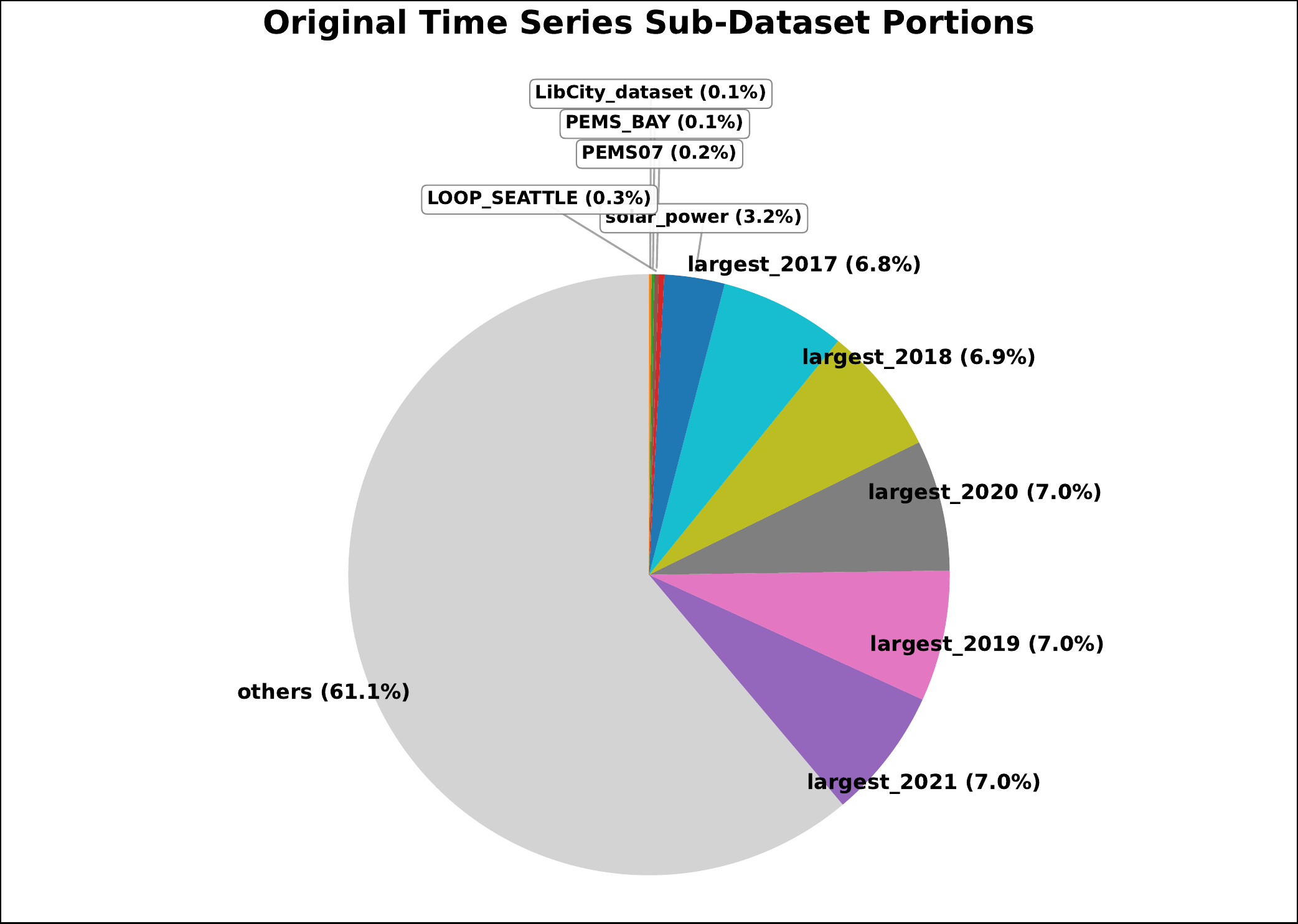} 
        \caption{Sub-dataset portion in training dataset.}
        \label{fig:sub2}
    \end{subfigure}
    \caption{Comparison between sub-dataset data contribution and data portion.}
    \label{fig:case-study-1}
\end{figure}

\paragraph{Contribution of each sub-dataset.}
In the first case study, we present the per-sub-dataset TSIS, i.e., $\Phi_{\mathcal{D}_l}$, throughout the training process (Figure~\ref{fig:case1-per-subset-TSIS}), along with the original proportion of each sub-dataset in the training corpus (Figure~\ref{fig:sub2}). The results reveal two key insights: (1) the contribution of each sub-dataset evolves during training, which intuitively motivates the use of online data augmentation; and (2) the contribution of a training sub-dataset does not necessarily align with its size. For instance, a relatively small sub-dataset (e.g., \texttt{solar\_power}) can contribute substantially to reducing test loss, whereas some large sub-datasets (e.g., those grouped under \texttt{others}, which account for 60\% of the total data in Figure~\ref{fig:sub2}) provide only a limited contribution (around 10\% in Figure~\ref{fig:case1-per-subset-TSIS}). Such findings are difficult to uncover through heuristic design or empirical intuition alone, but could be revealed through principled influence analysis.

\paragraph{Selected high-quality prompts and generated samples.}

\begin{figure}[h]
    \centering
    \includegraphics[width=\linewidth]{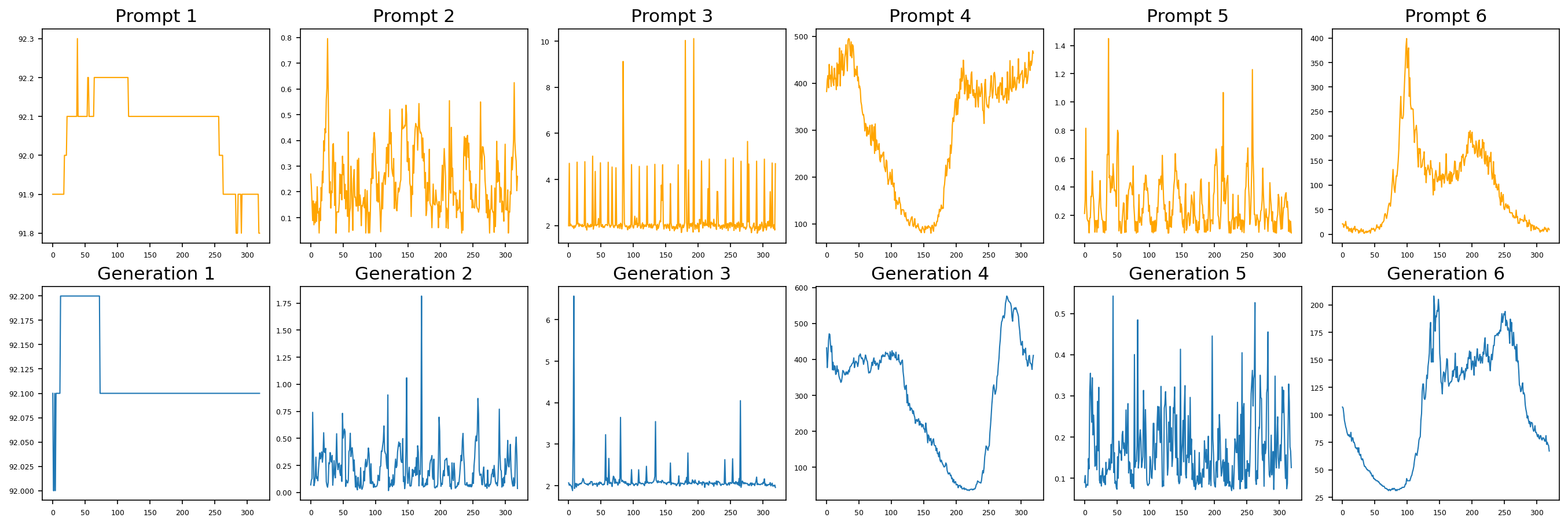}
    \caption{(\textbf{First row}) High-quality data samples prompt. (\textbf{Second row}) The corresponding guided generation.}
    \label{fig:case-study-2}
\end{figure}

In the second case study, we present the high-quality data samples prompts selected according to TSIS, and the corresponding guided generation in Figure~\ref{fig:case-study-2}. The results show that the generation model of the high-quality guided data augmentation could generate realistic data samples similar to the prompt with various patterns.

\section{Conclusion}
In this paper, we introduce \texttt{OATS}, a principled strategy for generating synthetic data tailored to different training steps in TSFMs. The core idea behind \texttt{OATS} is to leverage valuable training samples as principled guiding signals and dynamically generate high-quality synthetic data conditioned on them. To further improve efficacy and efficiency, we propose an explore–exploit mechanism that leverages cached influence scores to reduce computational overhead. Empirical evaluations across diverse model architectures, datasets, and prediction lengths demonstrate that \texttt{OATS} substantially outperforms both standard data augmentation methods and regular training in terms of TSFM performance. By expanding the scope of online TSFM data augmentation, \texttt{OATS} enables researchers to systematically optimize TSFM training datasets in a principled way.


\newpage
\appendix
\section{Related Works}
\paragraph{Data Augmentation in TSFMs.} Various data augmentation methods have been proposed in TSFM studies to enrich training datasets with realistic and diverse synthetic data. These approaches can be broadly categorized into two groups. The first category directly incorporates \emph{manually designed patterns} to synthesize training samples. For instance, Moment~\citep{goswami2024moment} introduces sinusoidal waves with varying frequencies, while ForecastPFN~\citep{10.5555/3666122.3666234} employs time series decomposition to generate components such as trend and seasonality using hand-selected hyperparameters. Similarly, TempoPFN~\citep{moroshan2025tempopfnsyntheticpretraininglinear} transforms synthetic data during pre-training to optimize linear recurrent neural networks. To address task-specific needs, TimeRCD~\citep{lan2025foundationmodelszeroshottime} utilizes a synthetic engine to inject anomalies for zero-shot detection, and TimesFM~\citep{das2024decoder} integrates ARMA processes to ensure the synthetic data remains sufficiently complex. Other works~\citep{shi2024time, ansari2024chronos} leverage KernelSynth to produce ``high-quality'' samples from handcrafted kernel banks, a strategy that CauKer~\citep{xie2025caukerclassificationtimeseries} later validated for pre-training classification-based TSFMs.

Alternatively, another group of methods applies \emph{basic transformations to existing time series} to generate new samples. For example, smoothing and jittering~\citep{um2017data} are used to enhance diversity through noise injection, while \citet{ansari2024chronos} proposes TSMixup to synthesize samples via weighted summations. Furthermore, Chronos-2~\citep{ansari2025chronos2univariateuniversalforecasting} creates multivariate series by imposing dependencies across univariate synthetic datasets. While effective, these approaches often rely on carefully crafted heuristics determined before training, leaving open the question of how to identify optimal augmentation strategies in a principled manner. Notably, although \citet{tagafilter} proposes an online algorithm combining RHO-based selection with traditional transformations, its simplistic generation process may lead to performance degradation. Moreover, the RHO-based approach incurs substantial computational overhead as it requires training a separate reference model.

\paragraph{Training Data Attribution.}  
Training data attribution aims to quantitatively estimate the influence of each training example on model behavior. This research direction traces back to robust statistics in the 1980s and has recently gained renewed attention with the use of influence functions to interpret black-box neural networks~\citep{koh2017understanding}. More recent work has focused on improving both the accuracy~\citep{wang2024capturing,ilyas2022datamodels} and scalability~\citep{park2023trak,choe2024your,grosse2023studying} of attribution methods. These approaches have also been applied in practical scenarios such as data selection~\citep{wang2024greats} and data valuation for data markets~\citep{deng2023computational}.

\paragraph{Time Series Generation.}  
Time series generation has been explored using a variety of model architectures. GAN-based approaches train adversarial objectives to generate realistic sequences~\citep{yoon2019time}. VAE-based methods design specialized decoders tailored for time series~\citep{desai2021timevae}. More recently, diffusion-based methods have been developed, such as causal time series generation~\citep{xia2025causal} and continuous time series generation~\citep{zhang2026mn}. Beyond unconditional synthesis, conditional generation has been studied in settings such as metadata-conditioned generation~\citep{narasimhan2024time} and prompt-based generation~\citep{huang2025timedp}.

\section{Complexity Analysis}\label{appendix:algorithm}

\texttt{OATS} involves computational overhead to the training process, which can be spitted into two sources: 1) calculation of TSIS and 2) synthetic data generation. \textbf{In summary}, the overhead of TSIS calculation is up to $\epsilon\times$ of the regular training step. The good performance shown in Figure~\ref{fig:epsilon} with $\epsilon=0.3$ shows that this overhead could be relatively small. The overhead of synthetic data generation model is a fixed number, which gets lower as the regular training step has higher time cost, e.g., a larger TSFM with more parameters.

\paragraph{Calculation of TSIS}

For TSIS calculation, the overhead of Equation~\ref{eq:itqs} mainly comes from the gradient calculation of reference data samples in ``explore'' steps. In our experiment, we use a small number of reference data samples (32) that matches the training batch size. This helps maintain the same computation complexity between TSIS and the regular gradient descent training step as \(\mathcal{O}(b)\), where $b$ is the training batch size. Additionally, we also adopted an efficient implementation called ``Ghost Inner Product''~\citep{wang2024data} to avoid per-sample gradient calculation while get the dot product result.

\begin{lemma}[Ghost Inner Product]
In the influence function calculation in TSIS, which includes sample-level gradient dot products, i.e., $\nabla_w\ell(z^{(1)}, w_t) \cdot \nabla_w\ell(z^{(2)}, w_t)$. We may only perform the backpropagation once and calculate the gradient dot product. Ghost Inner Product perform the product layer by layer in a model and here we present the method for a simple linear layer.
\begin{align}\label{eq:ghost-inner-prod}
    \frac{\partial\ell^{(1)}}{\partial w} \cdot \frac{\partial\ell^{(2)}}{\partial w} = (a^{(1)} \otimes \frac{\partial\ell^{(1)}}{\partial s^{(1)}})\cdot(a^{(2)} \otimes \frac{\partial\ell^{(2)}}{\partial s^{(2)}}) = ((a^{(1)})^{\top}a^{(2)})((\frac{\partial\ell^{(1)}}{\partial s^{(1)}})^\top\frac{\partial\ell^{(2)}}{\partial s^{(2)}}),
\end{align}
where $a^{(1)}$ and $a^{(2)}$ is the linear layer's input,  $s^{(1)}$ and $s^{(2)}$ is the pre-activation output, and $\otimes$ indicates the outer product.
\end{lemma}

In TSIS, $\ell^{(1)}=\ell(z_i, w_t)$ in each training step $t$ and $z_i\in\mathcal{B}_t$ and $\ell^{(2)}=\ell(\mathcal{D}_{ref}, w_t)$. ``Ghost Inner Product'' enable the implementation of influence score in TSIS to carry out \textbf{only one additional backpropagation} on $\ell(\mathcal{B}_t, w_t) + \ell(\mathcal{D}_{ref}, w_t)$ to get all terms in Equation~\ref{eq:ghost-inner-prod} compared to regular training.

Furthermore, the overhead of ``exploit'' steps is small enough to be ignored since TSIS in Equation~\ref{eq:itqs} is not used in those steps, \textbf{which means that the computational overhead of TSIS will be reduced by (1-$\epsilon$) and memory footprint will be on par with the regular training}. In Figure~\ref{fig:epsilon}, $\epsilon=0.3$ also substantially outperform regular training process, which shows good potential of reducing overhead. Following table~\ref{table:tsis-overhead} is a time cost record for training batch size 32. It is worth noting that the actual implementation may affect the time cost.

\begin{table}[h]
\centering
\caption{Per-training step time cost under different settings of Explore-exploit ratio $\epsilon$.}\label{table:tsis-overhead}
\resizebox{0.7\textwidth}{!}{%
\begin{tabular}{@{}llllll@{}}
\toprule
$\epsilon$ & Regular & 0.3  & 0.5  & 0.7  & 1.0  \\ \midrule
Avg. time cost per step (s) on A40 GPU    & 0.74    & 0.95 & 1.09 & 1.20 & 1.32 \\ \bottomrule
\end{tabular}%
}
\end{table}

\paragraph{Synthetic data generation.}

For synthetic data generation through diffusion model sampling, it is hard to compare the complexity with the regular gradient descent training step. Empirically, we find that some accelerated sampling strategies like DDIM~\citep{ddim} could perform well enough with very small sample steps. Moreover, the geneartion overhead will be a fixed term and the relative proportion will get lower as the TSFMs get larger, e.g., with more parameters. The overhead runtime of current synthetic data generation is 0.022s per synthetic sample, which is $0.022 \times 32 \times 0.5 = 0.35$s additional to the Table~\ref{table:tsis-overhead}.

\paragraph{Empirical results at same training time.} We also show the performance of \texttt{OATS} and baselines' performance under the \textbf{same training time as regular training}. \texttt{OATS} still outperform other baselines. It is also worth noting that with a limited training time, \texttt{OATS} may be more far away from converging.
\begin{table}[h]
\centering
\caption{Performance (NLL) of \texttt{OATS} and baselines' performance under the same time cost.}
\resizebox{\textwidth}{!}{%
\begin{tabular}{@{}llllllll@{}}
\toprule
Dataset     & \texttt{OATS}($\epsilon$=0.3) & \texttt{OATS}($\epsilon$=0.5) & \texttt{OATS}($\epsilon$=0.7) & \texttt{OATS}($\epsilon$=1) & TSMixup & Jitter & Regular \\ \midrule
Electricity & \textbf{6.00}     & 6.11              & 6.02              & 6.04            & 6.15    & 6.14   & 6.26    \\
ETTh1       & 1.82              & \textbf{1.81}     & 1.85              & 1.83            & 1.88    & 1.94   & 1.89    \\ \bottomrule
\end{tabular}%
}
\end{table}

\section{Experiment Settings}\label{app:exp-settings}

\paragraph{TSFM Models.} We conduct the experiment on the TSFM models with the same configuration as the encoder-only model with a 10M parameter size in \citet{yao2024towards}. The model is a modified version of Moirai~\citep{woo2024unified}, introduced by \citet{yao2024towards}, which incorporates patch embedding, rotary positional embedding, and a mixture of distributions to better adapt to time series forecasting while preserving extensibility.

\paragraph{TSFM Training Process.} We adopt a similar training setting as in \citet{yao2024towards} for the experiment. We utilize the AdamW optimizer with a batch size of 32 and a maximum learning rate of $10^{-3}$ with a linear warm-up of $10^4$ training steps, followed by cosine decay for the remaining $2 \times 10^4$ steps.

\paragraph{Datasets for TSFM.} We pretrain the TSFM on the modified (balanced domain sample, quality filtering) LOTSA-100M dataset in the pretraining stage provided by \citet{yao2024towards}. We take the native sub-dataset division as subsets in our experiment. These models are then evaluated on the out-of-distribution LSF dataset~\citep{wu2023timesnet}, using various prediction lengths (96, 192, 336) and the same preprocessing pipeline as in \citet{yao2024towards}. The detailed information of evaluation datasets is stated in Table~\ref{table:evaluation-datasets}.

\begin{table}[h]
\centering
\caption{Evaluation datasets and properties.}\label{table:evaluation-datasets}
\resizebox{0.6\textwidth}{!}{%
\begin{tabular}{@{}|l|l|c|c|@{}}
\toprule
\textbf{Dataset} & \textbf{Domain} & \textbf{Frequency} & \textbf{\# Prediction Length} \\ \midrule
ETTh1            & Energy          & H                  & 96/192/336/720                           \\ \midrule
ETTh2            & Energy          & H                  & 96/192/336/720                           \\ \midrule
ETTm1            & Energy          & 15min              & 96/192/336/720                           \\ \midrule
ETTm2            & Energy          & 15min              & 96/192/336/720                           \\ \midrule
Electricity      & Energy          & H                  & 96/192/336/720                           \\ \midrule
Weather          & Climate         & H                  & 96/192/336/720                           \\ \bottomrule
\end{tabular}%
}
\end{table}

\paragraph{Generation Model.} We leverage the architecture in \citet{huang2025timedp} as the backbone model. The denoising network of the diffusion model employed  adopts a U-Net architecture comprising 4 up/down sampling blocks with 8 attention heads and the dimension of each head is 64. An additional class embedding with 64 dimension is added to the output of each block.

\paragraph{Generation Model Training Process.} We train a Latent Diffusion Model for time series generation using 200 diffusion time steps with a linear noise schedule ranging from 0.0005 to 0.1. The model employs L1 loss as the training objective and a dropout probability of 0.5 to enable classifier-free guidance. We set the base learning rate to 0.001.

\paragraph{Datasets for Generation Model.} We train the generation model for high-quality guided data augmentation described in Section~\ref{sec:method} by a sampled dataset from the training dataset of TSFM. We sample 5\% of the dataset in 20 selected subsets (Table~\ref{table:datasets-diffusion}) in LOTSA-100M as the training set of the diffusion model. The generation length of the diffusion model is set to 320, and the diffusion step is set to 200. We use DDIM for the sampling process with 20 steps.

\begin{table}[h]
\caption{Training datasets of generation model.}\label{table:datasets-diffusion}
\centering
\resizebox{0.5\textwidth}{!}{%
\begin{tabular}{@{}|l|l|l|@{}}
\toprule
\textbf{Dataset}       & \textbf{Domain} & \textbf{Frequency} \\ \midrule
CMIP6                  & Climate         & 6H                 \\
ERA5                   & Climate         & H                  \\
CloudOpsTSF            & CloudOTS        & 5T                 \\
Azure VM Traces 2017   & CloudOTS        & 5T                 \\
Loop Seattle           & Transport       & 5T                 \\
PEMS07                 & Transport       & 5T                 \\
PEMS Bay               & Transport       & 5T                 \\
Q-Traffic              & Transport       & 15T                \\
Largest 2017           & Transport       & 5T                 \\
Largest 2018           & Transport       & 5T                 \\
Largest 2019           & Transport       & 5T                 \\
Largest 2020           & Transport       & 5T                 \\
Largest 2021           & Transport       & 5T                 \\
Australian Electricity & Energy          & 30T                \\
Buildings900K          & Energy          & H                  \\
Solar Power            & Energy          & 4S                 \\
Favorita Sales         & Sales           & D                  \\
Wiki-Rolling           & Web             & D                  \\
LibCity                & Transport       & 5T                 \\
OthersLOTSA            & Energy          & H                  \\ \bottomrule
\end{tabular}%
}
\end{table}

\paragraph{Hyperparameters.} For experiments in Table~\ref{table:encoder_performance_overall} and Table~\ref{table:decoder_performance_overall}, we set hyperparameters of TSMixup baseline to have $k=2$ and $\lambda_0 \sim \mathcal{U}(0.1, 0.9)$, $\lambda_1  = 1 - \lambda_0$. For Jitter, we use $\sigma=0.03$. For \texttt{OATS}, we fixed $\epsilon=1$ in Table~\ref{table:encoder_performance_overall} and Table~\ref{table:decoder_performance_overall}. Experiments with $\epsilon < 1$ will be included in Figure~\ref{fig:epsilon}.

\section{Additional Results}\label{app:additional-results}
In Table~\ref{table:ablation}, we present an ablation study between \texttt{OATS} and a partially implemented ``\texttt{OATS} (Sel only)'' that only trains on the selected guiding signals. Essentially, ``\texttt{OATS} (Sel only)'' is Algorithm~\ref{alg:oats} without “Step 2” in both explore and exploit steps. The results shows that \texttt{OATS} outperform the partially implementation in all settings. The experiment is carried out on Encoder-only TSFM and ``ETT'' refers to ETTm2.

\begin{table}[h]
\centering
\caption{Ablation study of the high-quality data selection step.}\label{table:ablation}
\resizebox{0.75\textwidth}{!}{
\begin{tabular}{@{}l|l|llllll@{}}
\toprule
\multirow{3}{*}{Dataset} & \multirow{3}{*}{Pred. length} & \multicolumn{2}{l}{\multirow{2}{*}{\texttt{OATS}}} & \multicolumn{2}{l}{\multirow{2}{*}{\texttt{OATS} (Sel only)}} & \multicolumn{2}{l}{\multirow{2}{*}{Regular}} \\
                         &                               & \multicolumn{2}{l}{}                                & \multicolumn{2}{l}{}                                 & \multicolumn{2}{l}{}                          \\ \cmidrule(lr){3-8}
                         &                               & NLL                      & MAPE                     & NLL                       & MAPE                     & NLL               & MAPE                       \\ \midrule
\multirow{4}{*}{ETT} & 96 & \cellcolor{green!15}\textbf{1.879} & \cellcolor{green!15}\textbf{0.216} & \cellcolor{green!15}2.156 & \cellcolor{green!15}0.303 & 2.177 & 0.306 \\
 & 192 & \cellcolor{green!15}\textbf{1.872} & \cellcolor{green!15}\textbf{0.208} & \cellcolor{green!15}2.106 & \cellcolor{red!10}0.274 & 2.124 & 0.271 \\
 & 336 & \cellcolor{green!15}\textbf{1.839} & \cellcolor{green!15}\textbf{0.242} & \cellcolor{green!15}2.051 & \cellcolor{red!10}0.305 & 2.095 & 0.298 \\
 & 720 & \cellcolor{green!15}\textbf{1.938} & \cellcolor{green!15}\textbf{0.284} & \cellcolor{green!15}2.123 & \cellcolor{red!10}0.337 & 2.204 & 0.328 \\
\midrule
\multirow{4}{*}{Electricity}& 96 & \cellcolor{green!15}\textbf{5.882} & \cellcolor{green!15}\textbf{0.425} & \cellcolor{green!15}6.091 & \cellcolor{green!15}0.649 & 6.187 & 0.731 \\
 & 192 & \cellcolor{green!15}\textbf{5.967} & \cellcolor{green!15}\textbf{0.515} & \cellcolor{green!15}6.147 & \cellcolor{green!15}0.632 & 6.244 & 0.744 \\
 & 336 & \cellcolor{green!15}\textbf{5.985} & \cellcolor{green!15}\textbf{0.582} & \cellcolor{green!15}6.229 & \cellcolor{green!15}0.664 & 6.284 & 0.810 \\
 & 720 & \cellcolor{green!15}\textbf{6.053} & \cellcolor{green!15}\textbf{0.580} & \cellcolor{green!15}6.345 & \cellcolor{green!15}0.674 & 6.411 & 0.822 \\
\midrule
\multirow{4}{*}{Weather} & 96 & \cellcolor{green!15}\textbf{3.192} & \cellcolor{green!15}\textbf{1.498} & \cellcolor{green!15}3.456 & \cellcolor{green!15}2.305 & 3.522 & 2.347 \\
 & 192 & \cellcolor{green!15}\textbf{3.193} & \cellcolor{green!15}\textbf{1.778} & \cellcolor{green!15}3.464 & \cellcolor{red!10}2.661 & 3.485 & 2.303 \\
 & 336 & \cellcolor{green!15}\textbf{3.263} & \cellcolor{green!15}\textbf{1.702} & \cellcolor{green!15}3.536 & \cellcolor{red!10}2.630 & 3.582 & 2.125 \\
 & 720 & \cellcolor{green!15}\textbf{3.675} & \cellcolor{green!15}\textbf{1.889} & \cellcolor{green!15}3.916 & \cellcolor{red!10}2.638 & 4.016 & 1.991 \\
\bottomrule
\end{tabular}
}
\end{table}

\begin{figure}[h]
    \centering
    \includegraphics[width=\linewidth]{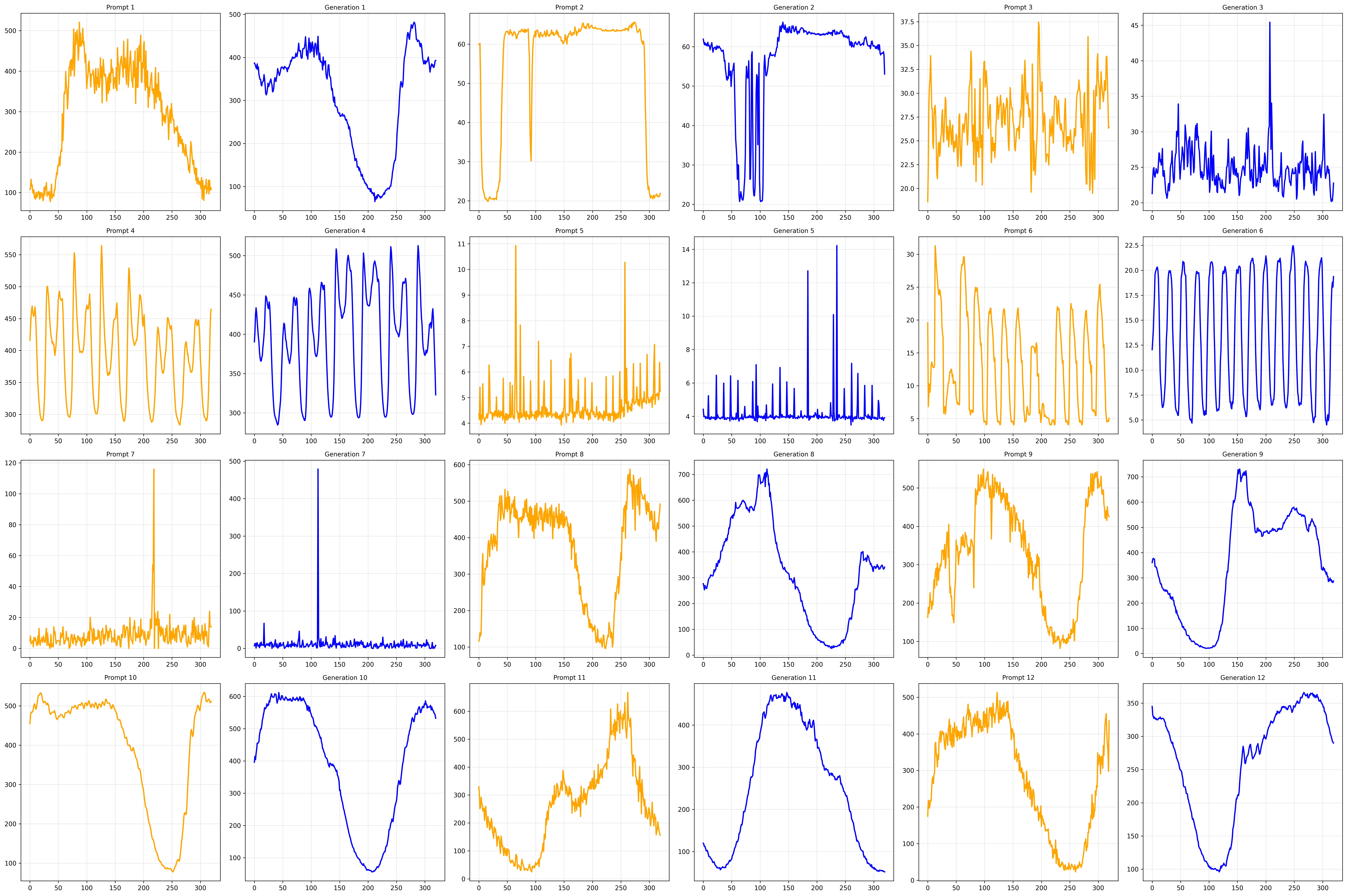}
    \caption{More examples of high-quality data samples prompts (yellow) and the corresponding guided generation (blue).}
    \label{fig:more-examples-generation}
\end{figure}

In Figure~\ref{fig:more-examples-generation}, we present more examples between high-quality data sample prompts and the corresponding generated samples.

\subsection{Improvement of class embedding on conditional generation.}
\begin{figure}[h]
    \centering
    \includegraphics[width=\linewidth]{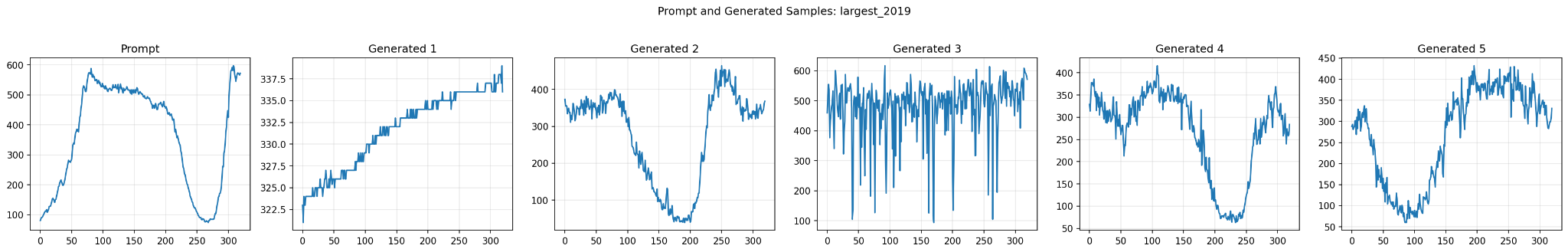}
    \includegraphics[width=\linewidth]{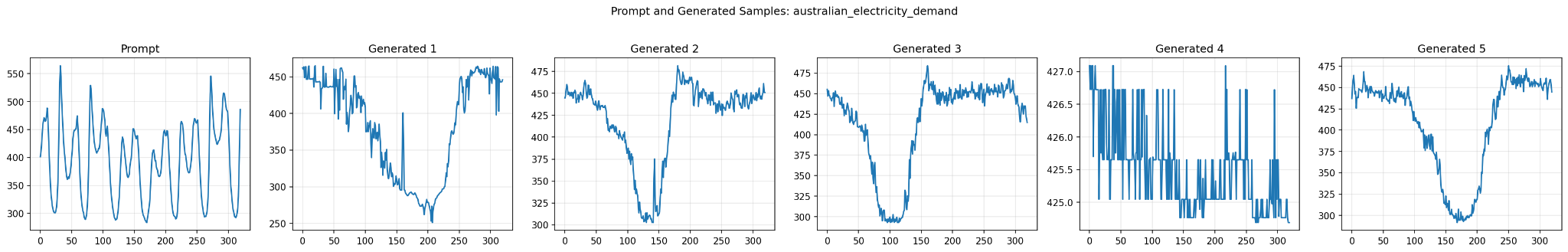}
    \includegraphics[width=\linewidth]{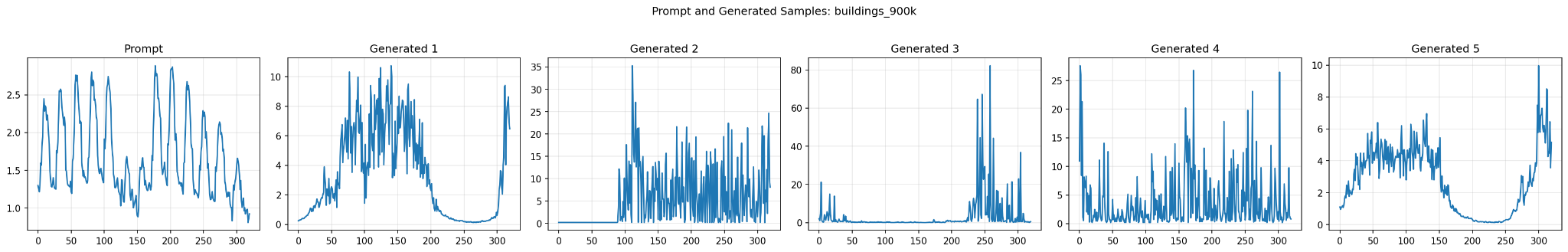}
    \caption{Examples of data generation of diffusion model \textbf{without} class embedding. For each row, the first sample is prompt and the following five are guided generations.}
    \label{fig:examples-generation-wo-class-condition}
\end{figure}

\begin{figure}[h]
    \centering
    \includegraphics[width=\linewidth]{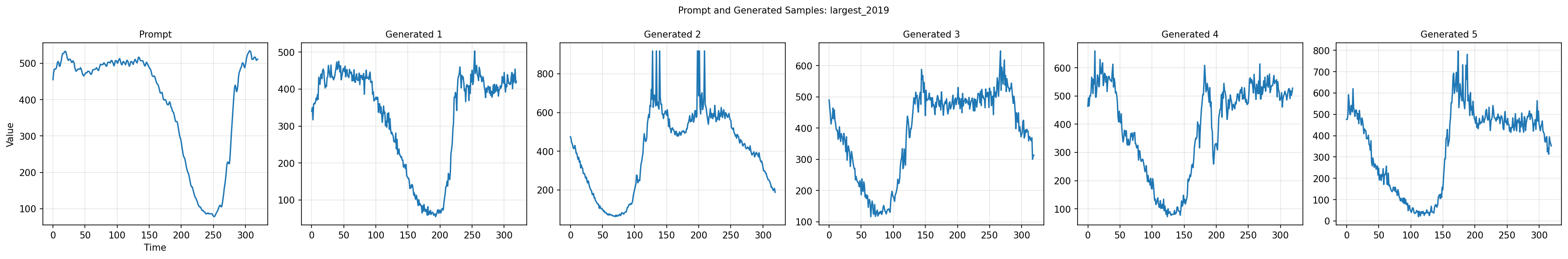}
    \includegraphics[width=\linewidth]{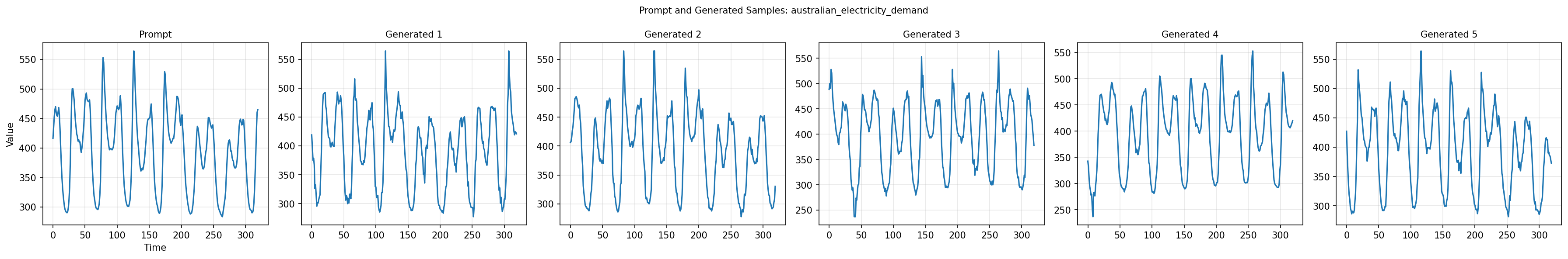}
    \includegraphics[width=\linewidth]{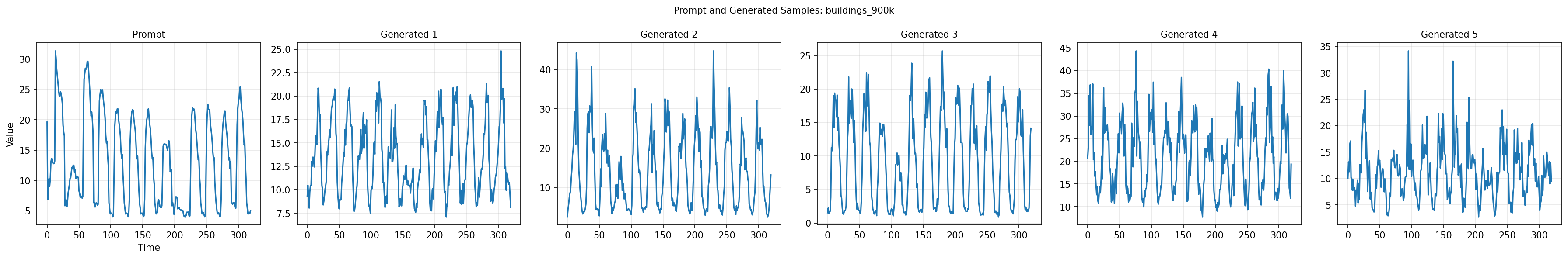}
    \caption{Examples of data generation of diffusion model \textbf{with} class embedding. For each row, the first sample is prompt, and the following five are guided generations.}
    \label{fig:examples-generation-w-class-condition}
\end{figure}

Figures~\ref{fig:examples-generation-wo-class-condition} and~\ref{fig:examples-generation-w-class-condition} demonstrate the improvement brought by class embeddings. In Figure~\ref{fig:examples-generation-wo-class-condition}, the samples generated without class embeddings (second to last in each row) exhibit broken patterns that fail to align with the prompts (first in each row). In contrast, Figure~\ref{fig:examples-generation-w-class-condition} illustrates that class embeddings help the model achieve better structural alignment with the prompts.

\subsection{Additional data-driven baseline}
We include an additional baseline using a data-driven generative model (a diffusion model, denoted as ``DD'') to generate synthetic data in an offline manner. The results in Table~\ref{table:new_baseline} show that OATS outperforms all baselines. The unconditional diffusion model is trained on the datasets listed in Table~\ref{table:datasets-diffusion}, incorporating 4 sampling blocks with 8 attention heads, following the same training procedure detailed in Appendix~\ref{app:exp-settings}.

\begin{table}[h]
\caption{Performance comparison of different methods with error bars. \textbf{Bold} means the best result. \colorbox{green!15}{Light green} background means that the performance is better than regular training process, while \colorbox{red!10}{light red} background means that the performance is worse than regular training process. The error bar shows the standard error of mean over 5 independent runs.}\label{table:new_baseline}
\resizebox{\textwidth}{!}{
\begin{tabular}{@{}l|l|llllllllll@{}}
\toprule
\multirow{3}{*}{Dataset} & \multirow{3}{*}{Pred. length} & \multicolumn{2}{l}{\multirow{2}{*}{OATS}} & \multicolumn{2}{l}{\multirow{2}{*}{DD}} & \multicolumn{2}{l}{\multirow{2}{*}{Jitter}} & \multicolumn{2}{l}{\multirow{2}{*}{TSMixUp}} & \multicolumn{2}{l}{\multirow{2}{*}{Regular}} \\
                         &                               & \multicolumn{2}{l}{}                        & \multicolumn{2}{l}{}                    & \multicolumn{2}{l}{}                   & \multicolumn{2}{l}{}                       & \multicolumn{2}{l}{}                       \\ \cmidrule(lr){3-12}
                         &                               & NLL                & MAPE                & NLL          & MAPE         & NLL          & MAPE         & NLL               & MAPE               & NLL               & MAPE               \\ \midrule
\multirow{4}{*}{ETTm1} & 96 & \cellcolor{green!15}\textbf{1.557 $\pm$ 0.030} & \cellcolor{green!15}\textbf{0.549 $\pm$ 0.029} & \cellcolor{green!15}1.764 $\pm$ 0.022 & \cellcolor{green!15}0.703 $\pm$ 0.007 & \cellcolor{green!15}1.721 $\pm$ 0.047 & \cellcolor{green!15}0.578 $\pm$ 0.014 & \cellcolor{green!15}1.658 $\pm$ 0.031 & \cellcolor{green!15}0.613 $\pm$ 0.031 & 1.823 $\pm$ 0.035 & 0.707 $\pm$ 0.074 \\
 & 192 & \cellcolor{green!15}\textbf{1.627 $\pm$ 0.042} & \cellcolor{green!15}\textbf{0.672 $\pm$ 0.043} & \cellcolor{green!15}1.833 $\pm$ 0.032 & \cellcolor{red!10}0.873 $\pm$ 0.015 & \cellcolor{green!15}1.715 $\pm$ 0.047 & \cellcolor{green!15}0.691 $\pm$ 0.044 & \cellcolor{green!15}1.725 $\pm$ 0.031 & \cellcolor{green!15}0.759 $\pm$ 0.038 & 1.870 $\pm$ 0.019 & 0.844 $\pm$ 0.056 \\
 & 336 & \cellcolor{green!15}\textbf{1.658 $\pm$ 0.032} & \cellcolor{green!15}\textbf{0.641 $\pm$ 0.048} & \cellcolor{green!15}1.838 $\pm$ 0.009 & \cellcolor{red!10}0.840 $\pm$ 0.011 & \cellcolor{green!15}1.763 $\pm$ 0.037 & \cellcolor{green!15}0.679 $\pm$ 0.035 & \cellcolor{green!15}1.765 $\pm$ 0.032 & \cellcolor{green!15}0.723 $\pm$ 0.031 & 1.870 $\pm$ 0.025 & 0.790 $\pm$ 0.059 \\
 & 720 & \cellcolor{green!15}\textbf{1.690 $\pm$ 0.029} & \cellcolor{green!15}\textbf{0.646 $\pm$ 0.050} & \cellcolor{green!15}1.848 $\pm$ 0.023 & \cellcolor{red!10}0.815 $\pm$ 0.026 & \cellcolor{green!15}1.809 $\pm$ 0.013 & \cellcolor{red!10}0.805 $\pm$ 0.017 & \cellcolor{green!15}1.787 $\pm$ 0.012 & \cellcolor{green!15}0.710 $\pm$ 0.026 & 1.869 $\pm$ 0.031 & 0.766 $\pm$ 0.052 \\
\bottomrule
\end{tabular}
}
\end{table}

\subsection{Sensitivity of initial partition}
In Table~\ref{table:granularity_comparison}, we follow the native partition structure (sub-datasets) provided in the LOTSA dataset and evaluate different settings of initial data granularity. ``Native Subdataset'' refers to the default setting used in the paper, where we maintain the original partition structure. ``Combine each two'' represents a coarser granularity setting where we merge two native subdatasets into a single partition. Additionally, we include a baseline with random partitioning, where data samples are randomly assigned to partitions while maintaining the same total number of partitions as the native subdataset configuration.

\begin{table}[h]
\caption{Performance comparison of different granularity partitions.}\label{table:granularity_comparison}
\centering
\resizebox{\textwidth}{!}{%
\begin{tabular}{l|l|llllll}
\hline
\multirow{3}{*}{Dataset} & \multirow{3}{*}{Pred. length} & \multicolumn{2}{l}{\multirow{2}{*}{Combine each two}} & \multicolumn{2}{l}{\multirow{2}{*}{Native Subdataset}} & \multicolumn{2}{l}{\multirow{2}{*}{Random Partition}} \\
                         &                               & \multicolumn{2}{l}{}                                  & \multicolumn{2}{l}{}                                   & \multicolumn{2}{l}{}                                  \\ \cline{3-8} 
                         &                               & NLL                       & MAPE                      & NLL                   & MAPE                           & NLL                       & MAPE                      \\ \hline
\multirow{4}{*}{ETTh1}   & 96                            & 1.837 $\pm$ 0.077         & 0.782 $\pm$ 0.029         & \textbf{1.760 $\pm$ 0.029}     & \textbf{0.694 $\pm$ 0.020}              & 1.965 $\pm$ 0.077         & 0.752 $\pm$ 0.029         \\
                         & 192                           & 1.860 $\pm$ 0.032         & 0.740 $\pm$ 0.047         & \textbf{1.766 $\pm$ 0.022}     & \textbf{0.682 $\pm$ 0.039}     & 1.969 $\pm$ 0.028         & 0.723 $\pm$ 0.041         \\
                         & 336                           & 1.885 $\pm$ 0.045         & 0.834 $\pm$ 0.045         & \textbf{1.825 $\pm$ 0.032}     & \textbf{0.746 $\pm$ 0.038}              & 1.998 $\pm$ 0.046         & 0.832 $\pm$ 0.033         \\
                         & 720                           & 1.911 $\pm$ 0.058         & 0.991 $\pm$ 0.063         & \textbf{1.853 $\pm$ 0.025}     & \textbf{0.947 $\pm$ 0.038}     & 2.009 $\pm$ 0.054         & 0.934 $\pm$ 0.048         \\ \hline
\end{tabular}%
}
\end{table}

We observe two key findings that align with our initial assumptions. First, we assume that the data exhibits locality within partitions based on semantic similarity (e.g., time series domains); consequently, random partitioning yields the poorest performance due to a lack of semantic coherence. Second, coarser granularity weakens the locality within each partition, thereby degrading performance relative to the ``Native Subdataset'' configuration.

\subsection{Sensitivity to decay factor $\beta$}
The performance in Table~\ref{table:beta_comparison} is not very sensitive to the $\beta$ setting as long as they are set to a reasonable range. In our experiment throughout the paper, we choose $\beta=0.01$.

\begin{table}[h]
\caption{Performance comparison of different beta values.}\label{table:beta_comparison}
\centering
\resizebox{\textwidth}{!}{%
\begin{tabular}{l|l|llllllll}
\hline
\multirow{3}{*}{Dataset} & \multirow{3}{*}{Pred. length} & \multicolumn{2}{l}{\multirow{2}{*}{$\beta$=0.1}} & \multicolumn{2}{l}{\multirow{2}{*}{$\beta$=0.01}} & \multicolumn{2}{l}{\multirow{2}{*}{$\beta$=0.001}} & \multicolumn{2}{l}{\multirow{2}{*}{$\beta$=0.0001}} \\
                         &                               & \multicolumn{2}{l}{}                             & \multicolumn{2}{l}{}                              & \multicolumn{2}{l}{}                               & \multicolumn{2}{l}{}                                \\ \cline{3-10} 
                         &                               & NLL                     & MAPE                   & NLL                 & MAPE                        & NLL                          & MAPE                & NLL                      & MAPE                     \\ \hline
\multirow{4}{*}{ETTh1}   & 96                            & 1.744 $\pm$ 0.066       & 0.814 $\pm$ 0.033      & 1.760 $\pm$ 0.029   & \textbf{0.694 $\pm$ 0.020}  & \textbf{1.688 $\pm$ 0.066}   & 0.758 $\pm$ 0.049   & 1.762 $\pm$ 0.078        & 0.739 $\pm$ 0.039        \\
                         & 192                           & 1.729 $\pm$ 0.041       & 0.727 $\pm$ 0.031      & 1.766 $\pm$ 0.022   & \textbf{0.682 $\pm$ 0.039}  & \textbf{1.717 $\pm$ 0.038}   & 0.714 $\pm$ 0.042   & 1.754 $\pm$ 0.038        & 0.704 $\pm$ 0.040        \\
                         & 336                           & 1.774 $\pm$ 0.036       & 0.835 $\pm$ 0.022      & 1.825 $\pm$ 0.032   & \textbf{0.746 $\pm$ 0.038}  & \textbf{1.753 $\pm$ 0.035}   & 0.783 $\pm$ 0.023   & 1.781 $\pm$ 0.039        & 0.787 $\pm$ 0.023        \\
                         & 720                           & 1.807 $\pm$ 0.054       & 1.061 $\pm$ 0.039      & 1.853 $\pm$ 0.025   & \textbf{0.947 $\pm$ 0.038}  & \textbf{1.783 $\pm$ 0.057}   & 0.962 $\pm$ 0.038   & 1.793 $\pm$ 0.056        & 0.953 $\pm$ 0.028        \\ \hline
\end{tabular}%
}
\end{table}

\subsection{Sensitivity to SNR filter bar}
We also conducted an experiment to evaluate the impact of the SNR threshold on performance in Table~\ref{table:snr_comparison}. This filtering mechanism is designed to eliminate noisy data points based on domain knowledge, thus complementing the influence score-based filtering. We observe that an overly strict threshold (e.g., 5dB) leads to the exclusion of excessive data, which degrades performance. Optimal results are achieved at relatively small values of $k$. While more sophisticated hyperparameter tuning strategies may exist, we leave such exploration for future work and adopt $k=3$dB for the remainder of this paper.

\begin{table}[h]
\caption{Performance comparison of different SNR values.}\label{table:snr_comparison}
\resizebox{\textwidth}{!}{
\begin{tabular}{@{}l|l|llllll@{}}
\toprule
\multirow{3}{*}{Dataset} & \multirow{3}{*}{Pred. length} & \multicolumn{2}{l}{\multirow{2}{*}{k=1dB}} & \multicolumn{2}{l}{\multirow{2}{*}{k=3dB}} & \multicolumn{2}{l}{\multirow{2}{*}{k=5dB}} \\
                         &                               & \multicolumn{2}{l}{}                        & \multicolumn{2}{l}{}                    & \multicolumn{2}{l}{}                       \\ \cmidrule(lr){3-8}
                         &                               & NLL                & MAPE                & NLL          & MAPE         & NLL               & MAPE               \\ \midrule
\multirow{4}{*}{ETTm1} & 96 & 1.630 $\pm$ 0.006 & 0.647 $\pm$ 0.012 & \textbf{1.557 $\pm$ 0.030} & \textbf{0.549 $\pm$ 0.029} & 1.672 $\pm$ 0.013 & 0.633 $\pm$ 0.013 \\
 & 192 & 1.709 $\pm$ 0.032 & 0.764 $\pm$ 0.017 & \textbf{1.627 $\pm$ 0.042} & \textbf{0.672 $\pm$ 0.043} & 1.753 $\pm$ 0.031 & 0.800 $\pm$ 0.022 \\
 & 336 & 1.751 $\pm$ 0.012 & 0.734 $\pm$ 0.015 & \textbf{1.658 $\pm$ 0.032} & \textbf{0.641 $\pm$ 0.048} & 1.769 $\pm$ 0.005 & 0.746 $\pm$ 0.022 \\
 & 720 & 1.785 $\pm$ 0.015 & 0.762 $\pm$ 0.017 & \textbf{1.690 $\pm$ 0.029} & \textbf{0.646 $\pm$ 0.050} & 1.775 $\pm$ 0.019 & 0.733 $\pm$ 0.026 \\
\bottomrule
\end{tabular}
}
\end{table}

\subsection{Sensitivity to reference set selection}
Table~\ref{table:batchsize_comparison} presents the results for various reference set sizes. To assess sensitivity, error bars are calculated based on randomly sampled reference sets. Throughout this paper, we use a fixed reference size of 32.
\begin{table}[h]
\caption{Performance comparison of different reference set sizes.}\label{table:batchsize_comparison}
\resizebox{\textwidth}{!}{
\begin{tabular}{@{}l|l|llllll@{}}
\toprule
\multirow{3}{*}{Dataset} & \multirow{3}{*}{Pred. length} & \multicolumn{2}{l}{\multirow{2}{*}{Ref size=8}} & \multicolumn{2}{l}{\multirow{2}{*}{Ref size=32}} & \multicolumn{2}{l}{\multirow{2}{*}{Ref size=128}} \\
                         &                               & \multicolumn{2}{l}{}                        & \multicolumn{2}{l}{}                    & \multicolumn{2}{l}{}                       \\ \cmidrule(lr){3-8}
                         &                               & NLL                & MAPE                & NLL          & MAPE         & NLL               & MAPE               \\ \midrule
\multirow{4}{*}{ETTm1} & 96 & 1.633 $\pm$ 0.012 & 0.515 $\pm$ 0.019 & 1.557 $\pm$ 0.030 & 0.549 $\pm$ 0.029 & \textbf{1.506 $\pm$ 0.005} & \textbf{0.505 $\pm$ 0.016} \\
 & 192 & 1.730 $\pm$ 0.030 & 0.728 $\pm$ 0.022 & 1.627 $\pm$ 0.042 & 0.672 $\pm$ 0.043 & \textbf{1.569 $\pm$ 0.022} & \textbf{0.584 $\pm$ 0.018} \\
 & 336 & 1.787 $\pm$ 0.007 & 0.758 $\pm$ 0.008 & 1.658 $\pm$ 0.032 & 0.641 $\pm$ 0.048 & \textbf{1.644 $\pm$ 0.013} & \textbf{0.587 $\pm$ 0.020} \\
 & 720 & 1.825 $\pm$ 0.020 & 0.807 $\pm$ 0.007 & 1.690 $\pm$ 0.029 & \textbf{0.646 $\pm$ 0.050} & \textbf{1.710 $\pm$ 0.022} & 0.672 $\pm$ 0.029 \\
\bottomrule
\end{tabular}
}
\end{table}

\end{document}